\documentclass[11pt]{article}

\usepackage[table,xcdraw]{xcolor}
\definecolor{NLcol}{HTML}{C04F15}
\definecolor{FOLcol}{HTML}{3B7D23}
\definecolor{MIXcol}{HTML}{1E77B4}
\definecolor{Finalcol}{HTML}{FF800E}
\definecolor{CMTgreen}{HTML}{228B22}
\definecolor{logiccol}{HTML}{FDE6D8}
\definecolor{affectcol}{HTML}{DDF2D8}
\definecolor{rhetoriccol}{HTML}{DCEBFA}
\usepackage{xcolor}
\usepackage{ragged2e}

\usepackage[most]{tcolorbox}

\newtcolorbox{findingbox}[1]{
  enhanced,
  breakable,
  colback=gray!4,
  colframe=gray!45!black,
  boxrule=0.4pt,
  arc=1.5pt,
  left=7pt,
  right=7pt,
  top=5pt,
  bottom=5pt,
  before skip=8pt,
  after skip=8pt,
  title=#1,
  coltitle=gray!35!black,
  fonttitle=\bfseries,
  attach title to upper,
  after title={\par\vspace{2pt}},
}
\usepackage[final]{acl}

\hypersetup{
    colorlinks=true,
    urlcolor=magenta,
    linkcolor=red
}

\usepackage{siunitx}
\usepackage{url}
\usepackage{amsthm}
\usepackage{booktabs}
\usepackage{multirow}
\usepackage{pifont}
\usepackage{listings}
\usepackage{xcolor}
\definecolor{codepurple}{rgb}{0.58,0,0.82}
\definecolor{codegray}{rgb}{0.5,0.5,0.5}
\definecolor{codeblue}{rgb}{0.0,0.2,0.6}
\definecolor{codered}{rgb}{0.8,0.0,0.0}
\definecolor{backcolour}{rgb}{0.98,0.98,0.98}
\usepackage[table]{xcolor}
\usepackage{multirow}
\usepackage{array}
\usepackage{tabularx}
\usepackage{diagbox}
\usepackage{algorithm}
\usepackage{algorithmicx}
\usepackage{algpseudocode}
\usepackage{makecell}
\usepackage[most]{tcolorbox}
\usepackage{todonotes}

\usepackage{enumitem}
\usepackage{amssymb} 

\usepackage{algpseudocode}
\usepackage{subcaption}
\usepackage{colortbl}

\usepackage{amsthm}

\theoremstyle{definition}

\theoremstyle{plain}

\usepackage{times}
\usepackage{latexsym}

\usepackage{CJKutf8}

\usepackage[T1]{fontenc}

\usepackage[utf8]{inputenc}

\usepackage{inconsolata}

\usepackage{graphicx}
\usetikzlibrary{positioning,arrows.meta,backgrounds,fit}
\usepackage{tikz}
\usetikzlibrary{shapes.geometric, arrows.meta, positioning, calc, chains, backgrounds, fit}
\usepackage{float} 

\definecolor{logiccol}{RGB}{232, 240, 254}
\definecolor{affectcol}{RGB}{252, 232, 230}
\definecolor{rhetoriccol}{RGB}{230, 244, 234}

\usepackage{graphicx}

\usepackage{graphicx}

\newcommand{\deepseek}{%
  \begin{tabular}[c]{@{}l@{}}
    \includegraphics[height=4ex]{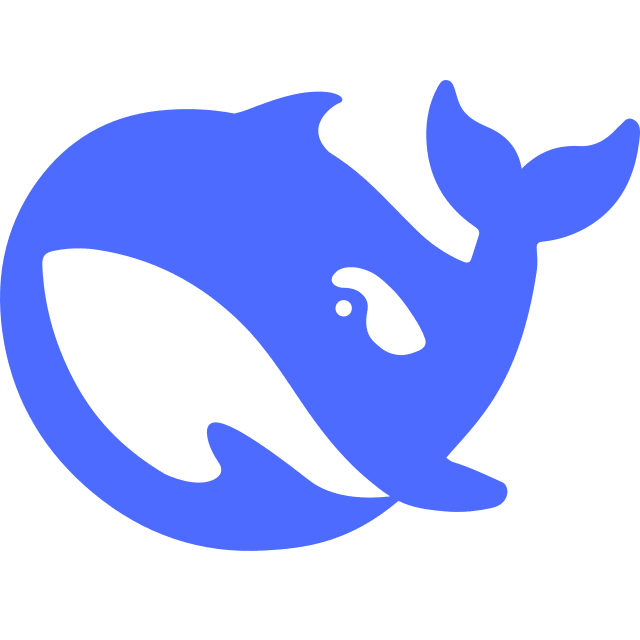} \\
  \end{tabular}%
}
\newcommand{\qwen}{%
  \begin{tabular}[c]{@{}l@{}}
    \includegraphics[height=4ex]{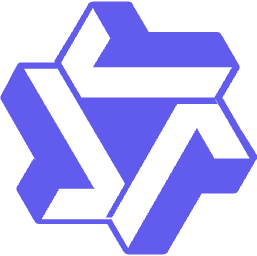} \\
  \end{tabular}%
}
\newcommand{\gpt}{%
  \begin{tabular}[c]{@{}l@{}}
    \includegraphics[height=4ex]{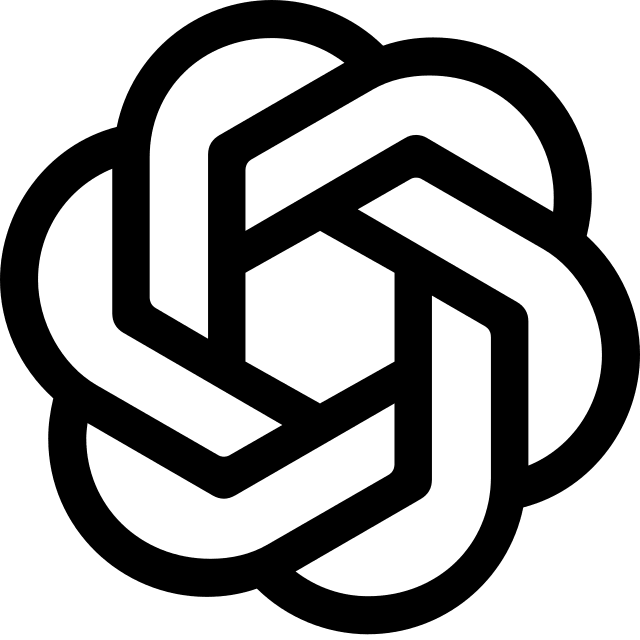} \\
  \end{tabular}%
}
\newcommand{\llama}{%
  \begin{tabular}[c]{@{}l@{}}
    \includegraphics[height=4ex]{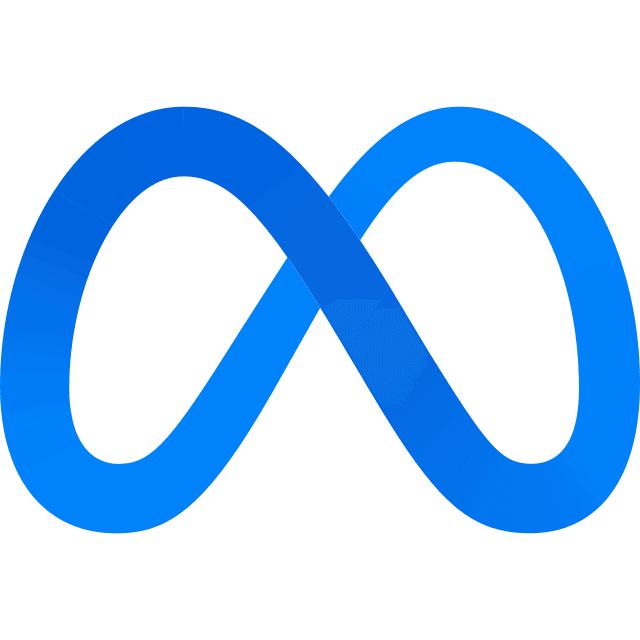} \\
  \end{tabular}%
}
\newcommand{\glm}{%
  \begin{tabular}[c]{@{}l@{}}
    \includegraphics[height=4ex,trim=0 53bp 0 53bp,clip]{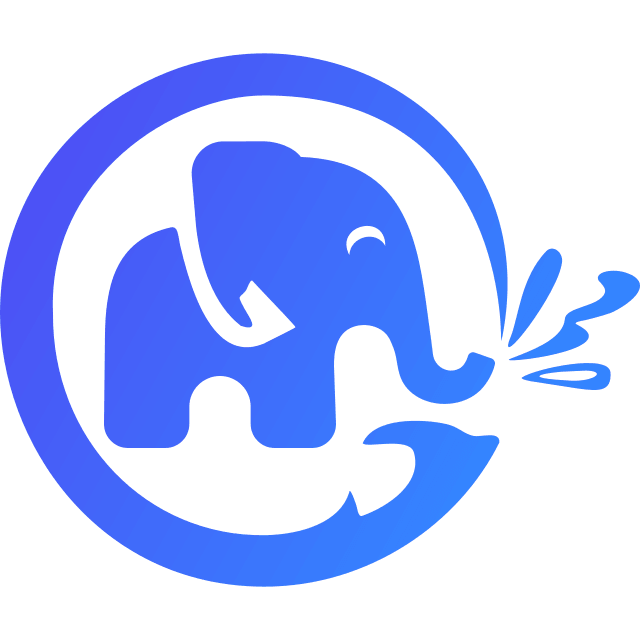} \\
  \end{tabular}%
}
\newcommand{\best}[1]{\textbf{#1}}
\newcommand{\second}[1]{\underline{#1}}
\newcommand{\posdelta}[1]{\textcolor{green!45!black}{#1}}
\newcommand{\negdelta}[1]{\textcolor{red!70!black}{#1}}
\usepackage{soul}
\sethlcolor{gray!20} 
\newenvironment{craddition}{\begingroup}{\endgroup}
\title{Surfacing the Unsaid: CUE-Bench for Affective Stance in Chinese Discourse}

\author{
 \textbf{Zhenyan Zheng\thanks{Equal contribution.}},
 \textbf{Yunyao Zhang\color{red}{\footnotemark[1]}},
 \textbf{Junxi Sheng},
 \textbf{Junqing Yu},
 \textbf{Zikai Song\thanks{Corresponding author.}}
\\
 Huazhong University of Science and Technology
\\
 \small{
   \{zhenyanzheng, ikostar, skyesong\}@hust.edu.cn
 }
}

\begin{document}

\maketitle

\begin{abstract}
Emotion understanding in discourse requires reasoning beyond surface sentiment, since speakers often convey affect through indirect, implicit, polite, ironic, or deliberately mismatched expressions.
Existing emotion benchmarks mainly annotate surface polarity or final emotion categories, while lacking a structured account of how explicit expression, implicit affect, pragmatic intent, and fine-grained emotion interact.
This limitation makes current evaluations insensitive to cases where affective meaning is concealed, weakened, inverted, or pragmatically reshaped, thereby obscuring models' failures in deeper emotion understanding.
To address this gap, we introduce \textbf{CUE-Bench}, a \textbf{C}hinese \textbf{U}nsaid \textbf{E}motion benchmark that centers on \textbf{Affective Stance} and covers diverse communicative scenarios.
CUE-Bench constructs nine human-interpretable affective stances from Explicit-Implicit polarity interaction and further provides intent and fine-grained emotion annotations for structured affective inference.
Experiments show that incorporating Affective Stance improves fine-grained emotion recognition by \textbf{3.1} percentage points and pragmatic intent detection by \textbf{8.1} percentage points over strong baselines. CUE-Bench is publicly available at \url{https://ai4ss.github.io/CUE-Bench/}

\end{abstract}

\begin{figure}[t]
    \centering
    \includegraphics[width=\linewidth]{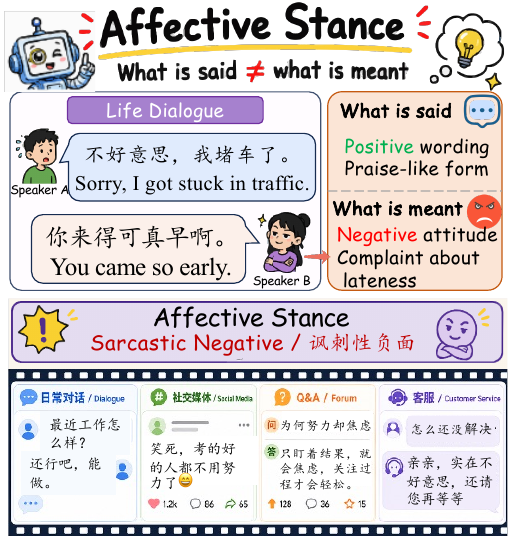}
    \caption{
    Overview of \textbf{CUE-Bench}. 
    The benchmark links \textit{what is said} and \textit{what is meant} through Affective Stance, enabling structured affective inference in Chinese discourse.
    }
    \label{fig1:teaser}
\end{figure}

\section{Introduction}

Understanding emotion in language is a central problem for affective computing and human-centered NLP~\cite{picard1997affective,goemotions-Fine-Grained-Emotions-acl2020,dai-etal-2026-psyche}.
However, real discourse often requires more than recognizing surface sentiment or predicting an isolated emotion category~\cite{ye2026omnitrend,intervenSim-2026}. 
It requires modeling how explicit expression and implicit affective tendency jointly shape the speaker's communicative orientation, which we define as \textit{Affective Stance}~\cite{du2007stance}. 
For example, positive wording may imply criticism, neutral wording may conceal commitment, and negative wording may signal affiliation or support. 
Such explicit-implicit mismatches are common in Chinese discourse, where affect is shaped by politeness, suppression, irony, understatement, and other indirect strategies~\cite{brown1987politeness,grice1975logic,gross1998emerging,dai-etal-2026-tears,dai2026dontboxin}. 
Without modeling this explicit-implicit relation, existing evaluations may capture either explicit expression or implicit affect in isolation, but miss how their mismatch shapes the speaker's affective stance.

Existing emotion benchmarks have advanced affective understanding~\cite{liao2026unlocking} from three perspectives. 
Representative resources include implicit emotion recognition benchmarks such as IEST~\cite{iest-Implicit-Emotions-acl2018}, SMP2020-EWECT~\cite{SMP2020-EWECT}, and ResEmo~\cite{resemo-socialMediaContent-emotion-aclfinding2024}, intent understanding benchmarks such as DailyDialog~\cite{dailydialog-Multi-turn-Dialogue-Dataset-AFNLP2017}, MELD~\cite{MELD-Multimodal-Emotional-Dialogue-ACL2019}, and CPED~\cite{CPED-PersonalizedEmotionalDialogue}, and fine-grained emotion benchmarks such as GoEmotions~\cite{goemotions-Fine-Grained-Emotions-acl2020} and CMMA~\cite{CMMA-multimodal-multiAffection-Nips2023}. 
However, most of them either focus on explicit affect or implicit affect separately, and their evaluation settings are usually restricted to a single task or communicative domain. 
Consequently, they provide limited diagnostic power for assessing whether models can infer affective stance, pragmatic intent, and fine-grained emotion when expressed affect and unsaid affect are misaligned.

To address this gap, we introduce \textbf{CUE-Bench}, a \textbf{C}hinese \textbf{U}nsaid \textbf{E}motion benchmark for affective understanding. 
To operationalize affective stance, we propose the \textbf{Explicit-Implicit Stance Matrix}, a structured framework that explicitly models the interaction between explicit expression and implicit affective tendency. 
By connecting what is expressed with what remains unsaid, this framework provides a unified perspective for analyzing affective stance, pragmatic intent, and fine-grained emotion. 
Built on this framework, CUE-Bench contains \textbf{51,823} annotated instances with four levels of supervision: explicit and implicit affective layers, nine Affective Stances, eight pragmatic intents, and twenty-five fine-grained emotions. 
It covers diverse Chinese discourse scenarios beyond dialogue, including open-domain conversation, social media comments, sarcasm-oriented text, customer-service interactions, and question-answering content~\cite{GAS32025-ga,liao2026role}.

CUE-Bench supports \textit{Affective Stance Recognition}, \textit{Pragmatic Intent Understanding}, and \textit{Fine-grained Emotion Classification}, spanning explicit-implicit stance interaction, pragmatic motivation, and final emotion interpretation. 
It is built via a hybrid pipeline of dual-model agreement, human adjudication, and bias-controlled LLM adjudication, using LLMs as a constrained aid rather than a replacement for human validation~\cite{text-annotation,LLMasJudge-nips2023}.

In summary, our contributions are:
\begin{itemize}[leftmargin=10pt, topsep=2pt, itemsep=0pt, label={$\bullet$}]
    \item We introduce \textbf{CUE-Bench}, a Chinese discourse benchmark that provides a unified setting for unsaid emotion understanding through three connected tasks: \textit{Affective Stance Recognition}, \textit{Pragmatic Intent Understanding}, and \textit{Fine-grained Emotion Classification}. 
    \item We propose the \textbf{Explicit-Implicit Stance Matrix}, a structured framework for modeling affective stance, pragmatic intent, and fine-grained emotion.
\end{itemize}

\newcommand{\cmark}{\textcolor{green!50!black}{\ding{51}}}
\newcommand{\xmark}{\textcolor{red!65!black}{\ding{55}}}

\begin{table*}[t]
\centering
\small
\setlength{\tabcolsep}{4pt}
\renewcommand{\arraystretch}{1.15}
\resizebox{\linewidth}{!}{
\begin{tabular}{lccccccccc}
\toprule
\textbf{Benchmark} 
& \textbf{Lang.} 
& \textbf{Scale} 
& \textbf{Multi-} 
& \textbf{Exp.} 
& \textbf{Imp.} 
& \textbf{Stance} 
& \textbf{Intent} 
& \textbf{Fine-grained} 
& \textbf{Multi-task} \\
& & & \textbf{domain} 
& \textbf{affect} 
& \textbf{affect} 
& & & \textbf{emotion} 
& \\
\midrule

DailyDialog~\cite{dailydialog-Multi-turn-Dialogue-Dataset-AFNLP2017} 
& En. & 13.1k 
& \xmark & \cmark & \xmark & \xmark & \cmark & \xmark & \cmark \\

IEST~\cite{iest-Implicit-Emotions-acl2018} 
& En. & 191.7k 
& \xmark & \xmark & \cmark & \xmark & \xmark & \xmark & \xmark \\

MELD~\cite{MELD-Multimodal-Emotional-Dialogue-ACL2019} 
& En. & 13.0k 
& \xmark & \cmark & \xmark & \xmark & \xmark & \xmark & \xmark \\

GoEmotions~\cite{goemotions-Fine-Grained-Emotions-acl2020} 
& En. & 58.0k 
& \xmark & \cmark & \xmark & \xmark & \xmark & \cmark & \xmark \\

SMP2020-EWECT~\cite{SMP2020-EWECT} 
& Zh. & 13.6k 
& \xmark & \cmark & \xmark & \xmark & \xmark & \xmark& \xmark \\

CPED~\cite{CPED-PersonalizedEmotionalDialogue} 
& Zh. & 12k 
& \xmark & \cmark & \xmark & \xmark & \cmark & \cmark & \cmark \\

CMMA~\cite{CMMA-multimodal-multiAffection-Nips2023} 
& Zh. & 21.8k 
& \xmark & \cmark & \xmark & \xmark & \cmark & \xmark & \cmark \\

ResEmo~\cite{resemo-socialMediaContent-emotion-aclfinding2024} 
& Zh. & 72.5k 
& \xmark & \cmark & \xmark & \xmark & \xmark & \cmark & \cmark \\
 
\midrule
\textbf{CUE-Bench} 
& Zh. & \textbf{51.8k} 
& \cmark & \cmark & \cmark & \cmark & \cmark & \cmark & \cmark \\
\bottomrule
\end{tabular}
}
\caption{Comparison with representative emotion-related benchmarks.
Scale is reported at the utterance/comment/instance level when available; for CPED and DailyDialog, we report dialogue-level scale following the original paper.
Exp. and Imp. affect denote explicit and implicit affective supervision.
Stance denotes whether the benchmark provides Explicit-Implicit affective stance labels.
Multi-domain indicates coverage of diverse communicative scenarios beyond a single dialogue or social-media domain.}
\label{tab:benchmark_comparison}
\end{table*}

\section{Related Work}
\label{sec:related_work}

\subsection*{Emotion Benchmarks}
Prior emotion-related benchmarks cover three related but largely separate lines of work.
\textit{(1) Implicit emotion recognition.} 
IEST~\cite{iest-Implicit-Emotions-acl2018} and Chinese resources such as SMP2020-EWECT~\cite{SMP2020-EWECT} and ResEmo~\cite{resemo-socialMediaContent-emotion-aclfinding2024} study emotions that are not directly expressed through emotion words. 
Yet they mainly formulate implicitness as emotion prediction, rather than as a structured relation between explicit expression and implicit affective tendency. 
\textit{(2) Pragmatic Intent Understanding.} 
DailyDialog~\cite{dailydialog-Multi-turn-Dialogue-Dataset-AFNLP2017}, Diplomat~\cite{Diplomat-Nips2023}, PUB~\cite{PUB-PragmaticsUnderstanding-aclfinding2024}, and recent work on social intelligence modeling~\cite{song2026socialintelligence} provide resources or perspectives for dialogue acts, situated pragmatic reasoning, pragmatic capability evaluation, and social interpretation.
However, they do not explicitly model how pragmatic intent interacts with explicit and implicit affect to reshape the affective meaning of an utterance. 
\textit{(3) Fine-grained Emotion Classification.} 
GoEmotions~\cite{goemotions-Fine-Grained-Emotions-acl2020} and CMMA~\cite{CMMA-multimodal-multiAffection-Nips2023} offer rich emotion or multi-affection labels, but still largely treat affective states as independent categories. 
Overall, existing benchmarks either evaluate explicit or implicit affect in isolation, or focus on a single task or communicative domain. 
They therefore provide limited diagnostic power for jointly assessing affective stance, pragmatic intent, and fine-grained emotion when what is expressed and what remains unsaid diverge.

\subsection*{Affective Modeling Methods}

Existing methods also address these three aspects separately. 
\textit{(1) Implicit emotion recognition.} 
Early methods~\cite{NTUA-SLP-IEST2018,Bert-naacl2019,maskBert} rely on neural transfer learning, recurrent encoders, attention mechanisms, and pre-trained language models to infer emotions that are not explicitly expressed~\cite{8INTENT}; however, they usually predict implicit affect directly without modeling its relation to explicit expression. 
\textit{(2) Pragmatic Intent Understanding.} 
Intent-oriented methods~\cite{JointBERT-intent-slot-2019,SLIM-multiIntent-2021,COSMIC-emnlp2020,logicAgent-2026,LPT-2026logical,dong-etal-2026-mitigating} model communicative goals through pre-trained encoders, joint intent-slot learning, dialogue context, or commonsense-enhanced conversational reasoning~\cite{1COMBINER,4TEMA}, but they often treat intent as an independent target rather than explaining how it reshapes affective meaning. 
\textit{(3) Fine-grained Emotion Classification.} 
Fine-grained emotion methods~\cite{goemotions-Fine-Grained-Emotions-acl2020,CMMA-multimodal-multiAffection-Nips2023,Bert-naacl2019} typically map text into rich emotion taxonomies with supervised classifiers, contextual encoders, or multi-label Transformer models~\cite{6OFFSET,15R3}, while leaving the intermediate links among explicit affect, implicit affect, and pragmatic intent underexplored. 
In contrast, our \textbf{Explicit-Implicit Stance Matrix} provides a structured intermediate framework that connects explicit expression, implicit affective tendency, pragmatic intent, and fine-grained emotion across the three tasks.

\begin{figure*}[t]
    \centering
    \includegraphics[width=\textwidth]{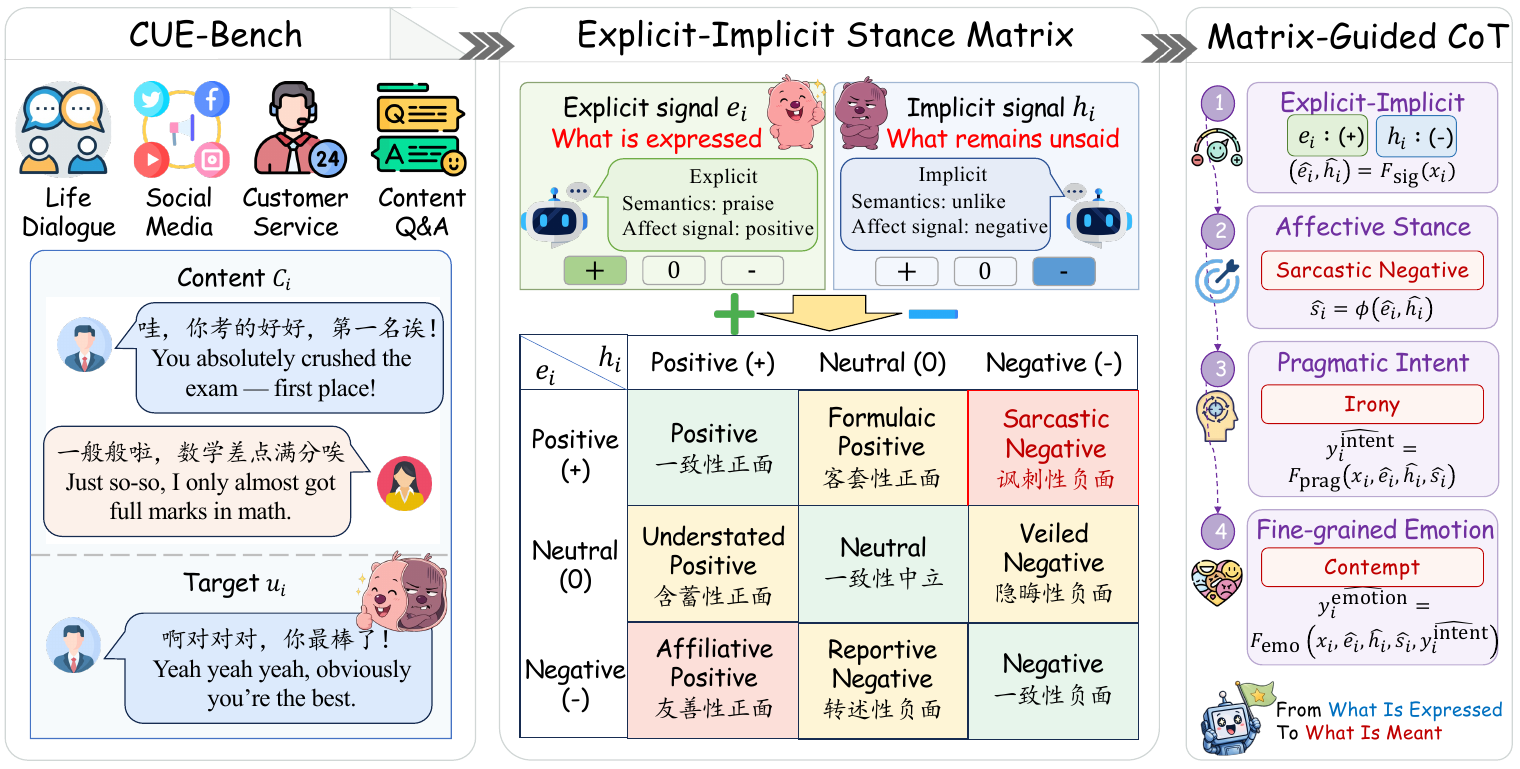}
    \caption{
    Overview of \textbf{CUE-Bench}. The benchmark collects context--target utterance pairs from diverse Chinese dialogue scenarios and models deeper affective understanding through the \textbf{Explicit-Implicit Stance Matrix}. The matrix contrasts the explicit affective signal $e_i$, i.e., what is expressed on the surface, with the implicit affective signal $h_i$, i.e., what remains unsaid. Guided by \textbf{Matrix-Guided CoT}, the reasoning pipeline progressively infers affective stance, pragmatic intent, and fine-grained emotion, moving from surface expression to intended meaning.
    }
    \label{fig:cuebench_pipeline}\end{figure*}

\section{CUE-Bench}
\label{sec:cue_bench}

\subsection{Design Principles}

CUE-Bench is designed to evaluate affective understanding beyond surface emotion classification, following three principles:

\begin{enumerate}[leftmargin=12pt, topsep=2pt, itemsep=0pt]
    \item \textbf{Context-sensitive affective inference.}
    Each instance includes contextual information, since implicit affect often cannot be inferred from an isolated utterance or text span alone.

    \item \textbf{Explicit-Implicit stance modeling.}
    The benchmark emphasizes how explicit expression and implicit affective tendency jointly shape the speaker's affective stance, covering realistic Chinese discourse phenomena such as politeness, suppression, irony, understatement, teasing, and indirect refusal.

    \item \textbf{Unified multi-task evaluation.}
    The annotation schema supports three connected tasks: \textit{Affective Stance Recognition}, \textit{Pragmatic Intent Understanding}, and \textit{Fine-grained Emotion Classification}, enabling evaluation from stance interaction to intent reasoning and final emotion interpretation.
\end{enumerate}

\subsection{Data Collection and Instance Format}

We construct CUE-Bench from diverse Chinese discourse sources, including open-domain conversations, social media comments, sarcasm-oriented text, customer-service interactions, and question-answering content.
Candidate instances are retained when contextual information plausibly changes affective interpretation, such as when the target text contains explicit affective markers, follows emotionally charged context, involves politeness or irony, or suggests a mismatch between literal wording and implicit affective tendency.

Each instance is represented as:
\[
x_i = (C_i, u_i),
\]
where $C_i$ denotes the discourse context, and $u_i$ denotes the target utterance or text span.

The annotation target is defined as:
\[
y_i = (y_i^{\mathrm{exp}}, y_i^{\mathrm{imp}}, y_i^{\mathrm{stance}}, y_i^{\mathrm{intent}}, y_i^{\mathrm{emotion}}),
\]
where $y_i^{\mathrm{exp}}$ and $y_i^{\mathrm{imp}}$ denote explicit and implicit affective layers, $y_i^{\mathrm{stance}}$ denotes the Affective Stance label, $y_i^{\mathrm{intent}}$ denotes pragmatic intent, and $y_i^{\mathrm{emotion}}$ denotes fine-grained emotion.

Before annotation, we normalize whitespace, remove duplicated instances, and discard samples that require private or external knowledge for interpretation. 
Personally identifiable information is masked, and sensitive content is retained only when it is necessary for affective interpretation and can be safely anonymized.

\subsection{Annotation Protocols}

Our annotation strategy combines model-assisted candidate generation, human adjudication, and bias-controlled LLM adjudication. 
The goal is to obtain scalable annotations while preserving human verification for ambiguous Explicit-Implicit affective relations.

\noindent\textbf{Model-assisted Candidate Annotation.}
We first use two independent models, $M_1$ and $M_2$, to generate candidate annotations for each instance:
\[
\hat{y}_i^{(1)} = M_1(x_i), \qquad \hat{y}_i^{(2)} = M_2(x_i).
\]
Consistent outputs are retained as high-confidence candidates:
\[
\mathcal{D}_{agr} = \{x_i \mid \hat{y}_i^{(1)} = \hat{y}_i^{(2)}\},
\]
while inconsistent outputs are treated as ambiguous cases for further adjudication:
\[
\mathcal{D}_{dis} = \{x_i \mid \hat{y}_i^{(1)} \neq \hat{y}_i^{(2)}\}.
\]

\noindent\textbf{Human Adjudication and Gold Verification.}
For a subset of $\mathcal{D}_{dis}$, trained annotators verify the candidate annotations by selecting the more appropriate label or revising the annotation when neither candidate is adequate. 
This produces a human-verified subset that serves as gold supervision and calibration data for later quality control.

\noindent\textbf{Bias-controlled LLM Adjudication.}
For the remaining ambiguous cases, we use LLMs as constrained adjudicators rather than unconstrained annotators. 
Each case is adjudicated twice with the candidate order reversed:
\[
b_i^{\mathrm{fwd}} = A(x_i,\hat{y}_i^{(1)},\hat{y}_i^{(2)}), 
b_i^{\mathrm{rev}} = A(x_i,\hat{y}_i^{(2)},\hat{y}_i^{(1)}),
\]
where $b_i^{\mathrm{fwd}},b_i^{\mathrm{rev}}\in\{1,2\}$ indicate the selected candidate under each order.
We map the reversed decision back to the canonical order as $\bar{b}_i^{\mathrm{rev}}=3-b_i^{\mathrm{rev}}$.
The retained annotation is:
\[
\tilde{y}_i =
\begin{cases}
\hat{y}_i^{(b_i^{\mathrm{fwd}})}, & \text{if } b_i^{\mathrm{fwd}}=\bar{b}_i^{\mathrm{rev}}, \\
\varnothing, & \text{otherwise}.
\end{cases}
\]
Only instances with $\tilde{y}_i\neq\varnothing$ are added to the retained LLM-adjudicated subset.
This consistency check reduces positional bias and filters out unstable adjudication cases.

\begin{table}[t]
\centering
\footnotesize
\setlength{\tabcolsep}{4pt}
\renewcommand{\arraystretch}{1.12}
\begin{tabular}{lccc}
\toprule
\textbf{Metric}
& \makecell[c]{\textbf{Affective}\\\textbf{Stance}}
& \makecell[c]{\textbf{Pragmatic}\\\textbf{Intent}}
& \makecell[c]{\textbf{Fine-grained}\\\textbf{Emotion}} \\
\midrule
\# Classes        & 9      & 8      & 25     \\
Krippendorff's $\alpha$ & 0.5197 & 0.3388 & 0.3146 \\
Majority Agr.    & 0.7933 & 0.6133 & 0.4400 \\
Avg. $\kappa$    & 0.5490 & 0.3922 & 0.3369 \\
Cond. Avg. $\kappa$ & --   & 0.7894 & 0.6689 \\
\bottomrule
\end{tabular}
\caption{
Inter-annotator agreement on 300 expert re-annotated instances.
Cond. Avg. $\kappa$ denotes average Cohen's kappa computed on instances with consistent Affective Stance.
Detailed construction statistics and pairwise agreement results are reported in Appendix~\ref{app:annotation_agreement}.
}
\label{tab:iaa_summary}
\end{table}

\subsection{Statistics and Analysis}
\label{sec:statistics_analysis}

\noindent\textbf{Dataset Composition.}
CUE-Bench is constructed from 60,000 candidate instances through a hybrid pipeline of model agreement, human verification, and bias-controlled LLM adjudication. 
The final benchmark contains 51,823 retained instances, consisting of high-confidence model-agreement annotations, a human-verified gold subset, and order-consistent LLM-adjudicated annotations. 
Detailed construction statistics are reported in Appendix~\ref{app:annotation_agreement}.

\begin{craddition}
\noindent\textbf{Human Audit of Two-model Agreement.}
To further validate the reliability of samples accepted through two-model agreement, we conduct a stratified human audit on this subset.
We sample 100 instances by source domain and Affective Stance, and randomly select instances within each stratum for expert review.
The expert audit finds that 92.0\% of the sampled two-model-agreement annotations are acceptable.
The remaining errors are concentrated in high-ambiguity boundary cases, especially strong sarcasm and stance reversal, which are also the cases emphasized in our human-verified gold subset construction.
\end{craddition}

\noindent\textbf{Adjudication Validation and Noise Estimate.}
We validate LLM adjudication on the human-verified gold subset, where the adjudicator achieves 89\% accuracy when selecting between two model-generated candidate annotations.
After forward--reverse consistency filtering, we estimate that approximately 1,600 retained instances may remain uncertain, yielding an overall estimated contamination rate of 3.1\% in the final dataset.
This indicates that constrained LLM adjudication introduces limited residual noise while substantially improving dataset coverage.

\begin{figure}[t]
    \centering
    \includegraphics[width=\columnwidth]{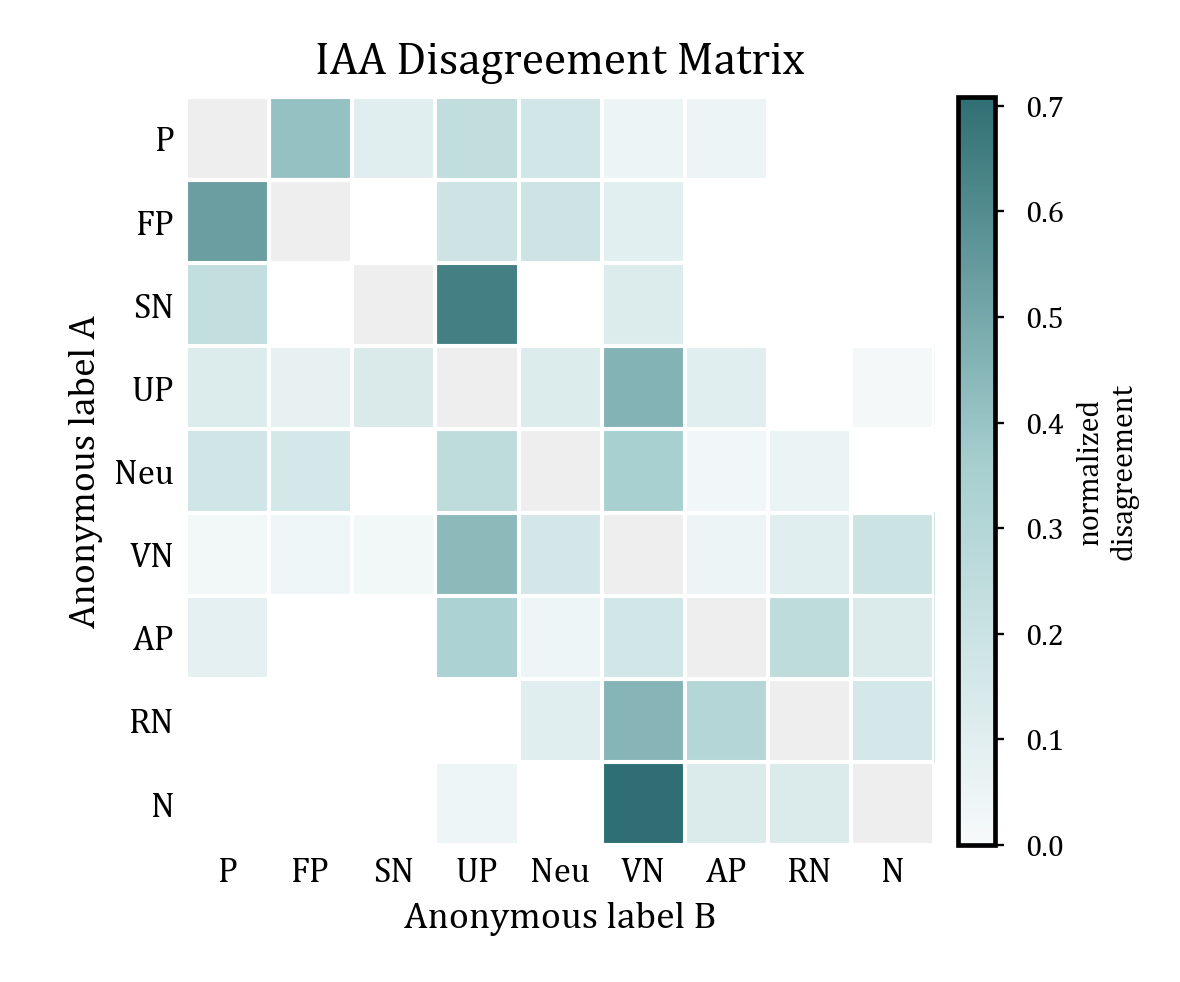}
    \caption{
    IAA disagreement matrix among three annotators.
    Darker cells indicate stronger disagreement between annotator label assignments.
    }
    \label{fig:annotation_iaa_matrix}
\end{figure}

\noindent\textbf{Inter-Annotator Agreement.}
To evaluate annotation reliability, we sample 300 instances from the human-verified gold subset and ask three expert annotators to independently re-annotate them.
Table~\ref{tab:iaa_summary} reports agreement at the three annotation layers.
Affective Stance obtains the strongest agreement, with a Krippendorff's $\alpha$ of 0.5197, a majority agreement rate of 79.3\%, and an average Cohen's $\kappa$ of 0.5490.
Pragmatic Intent and Fine-grained Emotion show lower raw agreement, with $\alpha$ scores of 0.3388 and 0.3146, respectively, reflecting the higher subjectivity of latent intent and fine-grained affective inference.
This pattern is consistent with prior findings that fixed IRR thresholds can be overly rigid for subjective annotation tasks, and that fine-grained emotion annotation often yields moderate or low chance-corrected agreement scores~\cite{Cross-replication-Reliability-acl2021, goemotions-Fine-Grained-Emotions-acl2020}.
We therefore report both raw and conditional agreement.
When conditioned on consistent Affective Stance, agreement improves substantially: the conditional average $\kappa$ reaches 0.7894 for Pragmatic Intent and 0.6689 for Fine-grained Emotion.
This suggests that Affective Stance provides a useful intermediate structure for localizing annotation ambiguity and reducing downstream uncertainty in intent and emotion labeling.
The annotators’ conflicts are not uniform across categories; the detailed confusion matrix is shown in Figure~\ref{fig:annotation_iaa_matrix}.

\section{Explicit-Implicit Stance Matrix}
\label{sec:EISM}

The core of CUE-Bench is the \textbf{Explicit-Implicit Stance Matrix}, a structured framework for modeling how expressed affect and unsaid affect jointly shape the speaker's affective stance. 
Rather than treating explicit and implicit affect as two independent labels, the matrix defines Affective Stance as their compositional relation.

\subsection{Explicit and Implicit Affective Signals}

For each instance $x_i=(C_i,u_i)$, we distinguish two affective signals. 
The \textit{explicit affective signal} describes the affect directly expressed by the target text $u_i$, such as affective words, intensifiers, punctuation, praise, complaint, apology, thanks, or direct evaluation. 
The \textit{implicit affective tendency} describes the affect inferred from the discourse context $C_i$ together with the target text $u_i$, including pragmatic force, speaker intention, and contextual implication.

Let $\pi(\cdot)$ denote an orientation projection that maps an affective layer
onto a three-way affective orientation space. Specifically, we define
\[
e_i=\pi(y_i^{\mathrm{exp}}), \qquad h_i=\pi(y_i^{\mathrm{imp}}),
\]
where $e_i,h_i\in\mathcal{O}$ and $\mathcal{O}=\{+,0,-\}$. The symbols $+$, $0$, and $-$ correspond to positive, neutral, and negative affective orientations, respectively. In this formulation, $e_i$ encodes the explicit affective signal anchored in the surface expression of $u_i$, while $h_i$ encodes the latent affective tendency inferred from the utterance together with its context $(C_i,u_i)$.

\definecolor{alignedbg}{HTML}{E6F4EA}
\definecolor{mediatedbg}{HTML}{E8F0FE}
\definecolor{reversedbg}{HTML}{FCE8E6}

\newcommand{\stanceone}[3]{%
\cellcolor{#1}\makecell[c]{\textsc{#2}\\[-1pt]\scriptsize $#3$}%
}

\newcommand{\stancetwo}[4]{%
\cellcolor{#1}\makecell[c]{\textsc{#2}\\[-1pt]\textsc{#3}\\[-1pt]\scriptsize $#4$}%
}






\subsection{Affective Stance}

We define Affective Stance as:
\[
s_i = \phi(e_i,h_i), \qquad s_i = y_i^{stance},
\]
where $\phi:\mathcal{O}\times\mathcal{O}\rightarrow\mathcal{S}$ maps each Explicit-Implicit pair to one of nine stance categories.

Table~\ref{fig:cuebench_pipeline} shows the resulting matrix, where rows correspond to the explicit signal $e_i$ and columns correspond to the implicit tendency $h_i$. 
This formulation makes Affective Stance a structured intermediate representation rather than an independent free-form label: if either $e_i$ or $h_i$ changes, the stance label $s_i$ changes accordingly. 
The matrix therefore provides a transparent mechanism for capturing alignment, neutralization, concealment, and reversal between what is expressed and what remains unsaid. 
Detailed definitions and examples of the nine Affective Stances are provided in Appendix~\ref{app:stance_definitions}.

\subsection{Matrix-Guided Chain-of-Thought}
\label{sec:matrix_guided_cot}

The Explicit-Implicit Stance Matrix further provides a structured chain-of-thought for affective inference. 
Rather than treating the three benchmark tasks as independent predictions, we use the matrix to impose an explicit reasoning order from surface expression to latent meaning and then to downstream affective interpretation. 
This turns affective understanding into a progressive inference process: identify what is expressed, infer what remains unsaid, resolve their stance relation, interpret the speaker's pragmatic motivation, and finally determine the fine-grained emotion.

For each instance $x_i$, the matrix-guided reasoning path is:
\[
\begin{aligned}
(\hat{e}_i,\hat{h}_i) 
&= F_{\mathrm{sig}}(x_i), \\
\hat{s}_i 
&= \phi(\hat{e}_i,\hat{h}_i), \\
\hat{y}_i^{\mathrm{intent}}
&= F_{\mathrm{prag}}(x_i,\hat{e}_i,\hat{h}_i,\hat{s}_i), \\
\hat{y}_i^{\mathrm{emotion}}
&= F_{\mathrm{emo}}(x_i,\hat{e}_i,\hat{h}_i,\hat{s}_i,\hat{y}_i^{\mathrm{intent}}).
\end{aligned}
\]
Here, $\hat{e}_i$ and $\hat{h}_i$ denote the predicted explicit and implicit affective orientations, $\hat{s}_i=\phi(\hat{e}_i,\hat{h}_i)$ denotes the predicted Affective Stance, $\hat{y}_i^{\mathrm{intent}}$ denotes the predicted pragmatic intent, and $\hat{y}_i^{\mathrm{emotion}}$ denotes the predicted fine-grained emotion.

This chain gives each prediction a structured dependency: stance is inferred from the Explicit-Implicit relation, intent is interpreted under the resulting stance, and fine-grained emotion is decided with both stance and intent as intermediate evidence. 
In LLM evaluation, we instantiate this process as a normalized prompting protocol, requiring the model to output intermediate fields in the order of explicit signal, implicit tendency, Affective Stance, pragmatic intent, and fine-grained emotion. 
Compared with direct label prediction, this matrix-guided CoT exposes the model's reasoning path and enables error analysis at each level of affective inference.

\begin{table*}[t]
\centering
\footnotesize
\setlength{\tabcolsep}{4.8pt}
\renewcommand{\arraystretch}{1.1}
\begin{tabular}{@{}p{0.11\textwidth}lcccccccccc@{}}
\toprule
\textbf{Model}
& \textbf{Method}
& \multicolumn{3}{c}{\textbf{Affective Stance} $\uparrow$}
& \multicolumn{3}{c}{\textbf{Pragmatic Intent} $\uparrow$}
& \multicolumn{3}{c}{\textbf{Fine-grained Emotion} $\uparrow$}
& \textbf{Avg.} $\uparrow$ \\
\cmidrule(lr){3-5}
\cmidrule(lr){6-8}
\cmidrule(lr){9-11}
&
& \textbf{Acc.}
& \textbf{F1}
& \textbf{W-F1}
& \textbf{Acc.}
& \textbf{F1}
& \textbf{W-F1}
& \textbf{Acc.}
& \textbf{F1}
& \textbf{W-F1}
& \\
\midrule

\multirow{4}{=}{\centering
\deepseek\\[-1pt]
\footnotesize DeepSeek\\[-2pt]
\footnotesize V4-Flash}
& Direct & 0.492 & 0.408 & 0.500 & 0.325 & 0.277 & 0.330 & 0.247 & 0.170 & 0.248 & 0.333 \\
& Few-shot & 0.480 & 0.412 & 0.498 & \second{0.339} & \second{0.288} & \second{0.344} & 0.247 & 0.174 & 0.246 & 0.336 \\
& CoT & \second{0.498} & \second{0.425} & \second{0.516} & 0.332 & 0.277 & 0.343 & \second{0.251} & \second{0.182} & \second{0.253} & \second{0.342} \\
   & \cellcolor{gray!15}\textbf{Ours} 
   & \cellcolor{gray!15}\best{0.510} 
   & \cellcolor{gray!15}\best{0.466} 
   & \cellcolor{gray!15}\best{0.527} 
   & \cellcolor{gray!15}\best{0.459} 
   & \cellcolor{gray!15}\best{0.361} 
   & \cellcolor{gray!15}\best{0.465} 
   & \cellcolor{gray!15}\best{0.313} 
   & \cellcolor{gray!15}\best{0.194} 
   & \cellcolor{gray!15}\best{0.327} 
   & \cellcolor{gray!15}\best{0.402} \\
& $\Delta$ & \posdelta{+0.012} & \posdelta{+0.041} & \posdelta{+0.011} & \posdelta{+0.120} & \posdelta{+0.073} & \posdelta{+0.121} & \posdelta{+0.062} & \posdelta{+0.012} & \posdelta{+0.074} & \posdelta{+0.061} \\

\midrule
\multirow{4}{=}{\centering
\gpt\\[-1pt]
\footnotesize GPT-4o\\[-2pt]
\footnotesize mini}
& Direct & 0.317 & 0.250 & 0.292 & 0.281 & 0.263 & 0.285 & 0.115 & 0.089 & 0.112 & 0.223 \\
& Few-shot & 0.341 & 0.281 & 0.336 & \second{0.302} & \best{0.274} & \second{0.298} & \second{0.226} & \best{0.153} & \second{0.229} & 0.271 \\
& CoT & \second{0.429} & \best{0.343} & \second{0.431} & 0.281 & 0.262 & 0.297 & \second{0.226} & \second{0.150} & 0.226 & \second{0.294} \\
   & \cellcolor{gray!15}\textbf{Ours} 
   & \cellcolor{gray!15}\best{0.458} 
   & \cellcolor{gray!15}\second{0.336} 
   & \cellcolor{gray!15}\best{0.444} 
   & \cellcolor{gray!15}\best{0.390} 
   & \cellcolor{gray!15}\second{0.264} 
   & \cellcolor{gray!15}\best{0.383} 
   & \cellcolor{gray!15}\best{0.275} 
   & \cellcolor{gray!15}0.147 
   & \cellcolor{gray!15}\best{0.266} 
   & \cellcolor{gray!15}\best{0.329} \\
& $\Delta$ & \posdelta{+0.029} & \negdelta{-0.007} & \posdelta{+0.013} & \posdelta{+0.088} & \posdelta{-0.010} & \posdelta{+0.085} & \posdelta{+0.049} & \negdelta{-0.006} & \posdelta{+0.037} & \posdelta{+0.035} \\
\midrule
\multirow{4}{=}{\centering
\llama\\[-1pt]
\footnotesize LLaMA-4\\[-2pt]
\footnotesize Maverick}
& Direct & 0.338 & 0.270 & 0.326 & 0.273 & 0.245 & 0.273 & 0.237 & 0.162 & 0.232 & 0.262 \\
& Few-shot & 0.367 & 0.315 & 0.385 & \second{0.293} & \second{0.271} & \second{0.307} & 0.233 & \second{0.169} & 0.233 & 0.286 \\
& CoT & \second{0.454} & \second{0.361} & \second{0.463} & 0.248 & 0.216 & 0.256 & \second{0.253} & 0.167 & \second{0.247} & \second{0.296} \\
   & \cellcolor{gray!12}\textbf{Ours} 
   & \cellcolor{gray!12}\best{0.512} 
   & \cellcolor{gray!12}\best{0.410} 
   & \cellcolor{gray!12}\best{0.511} 
   & \cellcolor{gray!12}\best{0.453} 
   & \cellcolor{gray!12}\best{0.342} 
   & \cellcolor{gray!12}\best{0.452} 
   & \cellcolor{gray!12}\best{0.339} 
   & \cellcolor{gray!12}\best{0.183} 
   & \cellcolor{gray!12}\best{0.331} 
   & \cellcolor{gray!12}\best{0.393} \\
& $\Delta$ & \posdelta{+0.058} & \posdelta{+0.049} & \posdelta{+0.048} & \posdelta{+0.160} & \posdelta{+0.071} & \posdelta{+0.145} & \posdelta{+0.086} & \posdelta{+0.014} & \posdelta{+0.084} & \posdelta{+0.096} \\
\midrule
\multirow{4}{=}{\centering
\llama\\[-1pt]
\footnotesize LLaMA 3.1\\[-2pt]
\footnotesize 8B}
& Direct & 0.203 & 0.194 & 0.221 & 0.232 & 0.208 & 0.233 & 0.142 & 0.086 & 0.151 & 0.186 \\
& Few-shot & \second{0.281} & \second{0.256} & 0.307 & \second{0.259} & \second{0.231} & \second{0.277} & \second{0.167} & \best{0.112} & \second{0.166} & \second{0.228} \\
& CoT & 0.277 & \second{0.256} & \second{0.311} & 0.226 & 0.216 & 0.248 & 0.129 & \second{0.101} & 0.135 & 0.211 \\
   & \cellcolor{gray!12}\textbf{Ours} 
   & \cellcolor{gray!12}\best{0.362} 
   & \cellcolor{gray!12}\best{0.293} 
   & \cellcolor{gray!12}\best{0.369} 
   & \cellcolor{gray!12}\best{0.285} 
   & \cellcolor{gray!12}\best{0.236} 
   & \cellcolor{gray!12}\best{0.292} 
   & \cellcolor{gray!12}\best{0.190} 
   & \cellcolor{gray!12}0.097 
   & \cellcolor{gray!12}\best{0.184} 
   & \cellcolor{gray!12}\best{0.256} \\
& $\Delta$ & \posdelta{+0.081} & \posdelta{+0.037} & \posdelta{+0.058} & \posdelta{+0.026} & \posdelta{+0.005} & \posdelta{+0.015} & \posdelta{+0.023} & \negdelta{-0.015} & \posdelta{+0.018} & \posdelta{+0.027} \\
\midrule
\multirow{4}{=}{\centering
\qwen\\[-1pt]
\footnotesize Qwen-3-8B}
& Direct & 0.180 & 0.154 & 0.172 & 0.196 & 0.210 & 0.242 & 0.209 & 0.128 & \second{0.225} & 0.191 \\
& Few-shot & 0.181 & 0.183 & 0.197 & 0.231 & 0.214 & 0.227 & \second{0.212} & 0.121 & 0.221 & 0.199 \\
& CoT & \second{0.422} & \best{0.325} & \best{0.430} & \second{0.254} & \second{0.226} & \second{0.257} & \second{0.212} & \second{0.132} & 0.206 & \second{0.274} \\
   & \cellcolor{gray!12}\textbf{Ours} 
   & \cellcolor{gray!12}\best{0.424} 
   & \cellcolor{gray!12}\second{0.309} 
   & \cellcolor{gray!12}\second{0.412} 
   & \cellcolor{gray!12}\best{0.364} 
   & \cellcolor{gray!12}\best{0.271} 
   & \cellcolor{gray!12}\best{0.367} 
   & \cellcolor{gray!12}\best{0.260} 
   & \cellcolor{gray!12}\best{0.137} 
   & \cellcolor{gray!12}\best{0.262} 
   & \cellcolor{gray!12}\best{0.312} \\
& $\Delta$ & \posdelta{+0.002} & \negdelta{-0.016} & \negdelta{-0.018} & \posdelta{+0.110} & \posdelta{+0.045} & \posdelta{+0.110} & \posdelta{+0.048} & \posdelta{+0.005} & \posdelta{+0.037} & \posdelta{+0.038} \\
\midrule
\multirow{4}{=}{\centering
\glm\\[-1pt]
\footnotesize GLM-5.1}
& Direct & 0.424 & 0.358 & 0.435 & 0.312 & 0.280 & 0.310 & \second{0.305} & 0.201 & 0.313 & \second{0.326} \\
& Few-shot & 0.417 & 0.361 & 0.424 & \second{0.322} & \second{0.295} & \second{0.327} & 0.284 & \second{0.203} & 0.299 & \second{0.326} \\
& CoT & \second{0.463} & \second{0.408} & \second{0.479} & 0.307 & 0.278 & \second{0.327} & 0.234 & 0.183 & 0.246 & 0.325 \\
   & \cellcolor{gray!12}\textbf{Ours} 
   & \cellcolor{gray!12}\best{0.549} 
   & \cellcolor{gray!12}\best{0.448} 
   & \cellcolor{gray!12}\best{0.552} 
   & \cellcolor{gray!12}\best{0.436} 
   & \cellcolor{gray!12}\best{0.357} 
   & \cellcolor{gray!12}\best{0.450} 
   & \cellcolor{gray!12}\best{0.333} 
   & \cellcolor{gray!12}0.185 
   & \cellcolor{gray!12}\best{0.333} 
   & \cellcolor{gray!12}\best{0.405} \\
& $\Delta$ & \posdelta{+0.086} & \posdelta{+0.040} & \posdelta{+0.073} & \posdelta{+0.114} & \posdelta{+0.062} & \posdelta{+0.123} & \posdelta{+0.028} & \negdelta{-0.018} & \posdelta{+0.020} & \posdelta{+0.079} \\
\bottomrule
\end{tabular}
\caption{
\textbf{Main results.}
We report Accuracy, macro-F1, and weighted-F1 on three CUE-Bench tasks: Affective Stance, Pragmatic Intent, and Fine-grained Emotion.
$\uparrow$ indicates that higher values are better.
}
\label{tab:main_results}
\end{table*}

\begin{craddition}
Importantly, this reasoning order should not be interpreted as a strict deterministic hierarchy.
Affective Stance, Pragmatic Intent, and Fine-grained Emotion are distinct annotation dimensions with structured dependencies.
Affective Stance captures the relation between surface expression and implicit affect; Pragmatic Intent describes the communicative function through which the utterance acts in context; Fine-grained Emotion preserves the more specific affective state.
Thus, the matrix provides an interpretable pragmatic scaffold for downstream inference, while intent and emotion remain separately annotated targets rather than labels mechanically determined by stance.
\end{craddition}

\section{Experiments}
\label{sec:experiments}

\subsection{Settings}

\paragraph{Models.}
We evaluate a diverse set of large language models covering both proprietary and open-source families. 
(1) For proprietary models, we include GPT-4o-mini and DeepSeek-V4-Flash, which represent widely used lightweight instruction-following models with strong general reasoning ability. 
(2) For open-source models, we evaluate LLaMA-4-Maverick, LLaMA-3.1-8B, Qwen-3-8B, and GLM-5.1. 

\paragraph{LLM prompting baselines.}
We compare our matrix-guided reasoning method with three standard LLM prompting baselines:
(1) \textbf{Direct prompting}, which asks the model to directly output the predicted label from the context and target utterance;
(2) \textbf{Few-shot prompting}, which provides annotated demonstrations before prediction;
and (3) \textbf{CoT prompting}, which elicits free-form intermediate reasoning before the final prediction.

\paragraph{Metrics.}
We report \textbf{Accuracy (Acc.)}, \textbf{macro-F1 (F1)}, and \textbf{weighted-F1 (W-F1)}. 
Accuracy measures overall correctness, macro-F1 gives equal weight to each class and reflects performance on minority categories, while weighted-F1 accounts for label imbalance by weighting class-wise F1 scores by class frequency.

\subsection{Main Results}

As shown in Table~\ref{tab:main_results}, we draw three observations.

\textbf{Our matrix-guided method achieves the best overall performance.}
Our method obtains the highest average score across all evaluated models, outperforming the strongest baseline by $+0.027$ to $+0.096$. 
The gains are especially clear on DeepSeek-V4-Flash, LLaMA-4-Maverick, and GLM-5.1, showing the effectiveness of jointly modeling surface expression and implicit affective tendency.

\textbf{Pragmatic Intent shows the most consistent gains.}
Our method improves Pragmatic Intent accuracy across all models, with gains from $+0.026$ to $+0.160$, and also consistently improves weighted-F1. 
This indicates that communicative intent benefits strongly from the Explicit-Implicit affective distinction.

\textbf{Fine-grained Emotion improves but remains harder.}
Our method improves Fine-grained Emotion accuracy and weighted-F1 across all models, while macro-F1 gains are less stable. 
This suggests that stance-guided reasoning helps infer emotional tendency, but distinguishing fine-grained emotion categories remains challenging.

\subsection{Ablations}
\label{sec:ablation}

\noindent\textbf{Oracle-conditioning ablation.}
To examine whether the proposed reasoning path benefits downstream affective inference, we conduct oracle-conditioning ablations with DeepSeek-V4-Flash and LLaMA-4-Maverick. 
As shown in Table~\ref{tab:ablation_oracle_chain}, we test whether gold Affective Stance and Pragmatic Intent can serve as intermediate evidence for predicting Pragmatic Intent and Fine-grained Emotion. 
We draw three observations.

\textbf{Affective Stance provides strong evidence for pragmatic intent.}
With gold Affective Stance, Pragmatic Intent prediction reaches 0.703 macro-F1 on DeepSeek-V4-Flash and 0.658 on LLaMA-4-Maverick, confirming its value as an intermediate representation.

\textbf{Pragmatic Intent adds complementary evidence for emotion inference.}
Adding gold Pragmatic Intent on top of gold Affective Stance improves Fine-grained Emotion accuracy and weighted-F1 for both models, showing that intent contributes information beyond stance alone.

\begin{table}[H]
\centering
\footnotesize
\setlength{\tabcolsep}{4pt}
\renewcommand{\arraystretch}{1.10}
\begin{tabular}{llccc}
\toprule
\textbf{Model} & \textbf{Task} & \textbf{Acc.} & \textbf{F1} & \textbf{W-F1} \\
\midrule
\multirow{3}{*}{DeepSeek-V4-Flash}
& I
& 0.791 & 0.703 & 0.790 \\
& II
& 0.392 & 0.243 & 0.379 \\
& III
& 0.432 & 0.219 & 0.389 \\
\midrule
\multirow{3}{*}{LLaMA-4-Maverick}
& I
& 0.739 & 0.658 & 0.745 \\
& II
& 0.430 & 0.251 & 0.436 \\
& III
& 0.489 & 0.254 & 0.466 \\
\bottomrule
\end{tabular}
\caption{
\textbf{Ablations on CUE-Bench}.
Task I: Affective Stance $\rightarrow$ Pragmatic Intent;
Task II: Affective Stance $\rightarrow$ Fine-grained Emotion;
Task III: Affective Stance + Pragmatic Intent $\rightarrow$ Fine-grained Emotion.
}
\label{tab:ablation_oracle_chain}
\end{table}

\textbf{Oracle signals help but do not replace category-level emotion discrimination.}
Although gold intermediate labels improve Fine-grained Emotion performance, the gains are not uniform across all metrics. 
This suggests that the reasoning chain provides useful affective evidence, while final emotion prediction still requires direct discrimination among emotion categories.

\subsection{Analysis}
\label{sec:analysis}

We focus on two research questions that explain the distributional and annotation patterns observed in the CUE-Bench.

\paragraph{\textbf{RQ1: Why are negative and indirect cases frequent?}} 
CUE-Bench contains many negative or negative-leaning cases because we intentionally retain sources rich in implicit affect, such as sarcastic, hostile, conflictual, and emotionally charged online discourse. This increases the density of cases where literal wording is insufficient, making the benchmark more diagnostic for unsaid affect.

\textbf{Veiled negative cases dominate the benchmark.}
\textsc{Veiled Negative} accounts for 22.3\% of the data, where neutral surface wording often implies dissatisfaction, reluctance, pressure, or indirect criticism. 
These cases are challenging because models must recover hidden negative affect from context rather than explicit lexical cues.

\textbf{Sarcastic negative cases reflect deliberate stress-test design.}
\textsc{Sarcastic Negative} also appears frequently (10.9\%) because the corpus includes sarcasm-oriented and hostile-comment data. 
Thus, the negative skew should be viewed as a benchmark feature rather than a natural base-rate estimate: CUE-Bench is designed to evaluate affective inference under pragmatic mismatch.

\begin{figure}[H]
    \centering
    \includegraphics[width=\columnwidth]{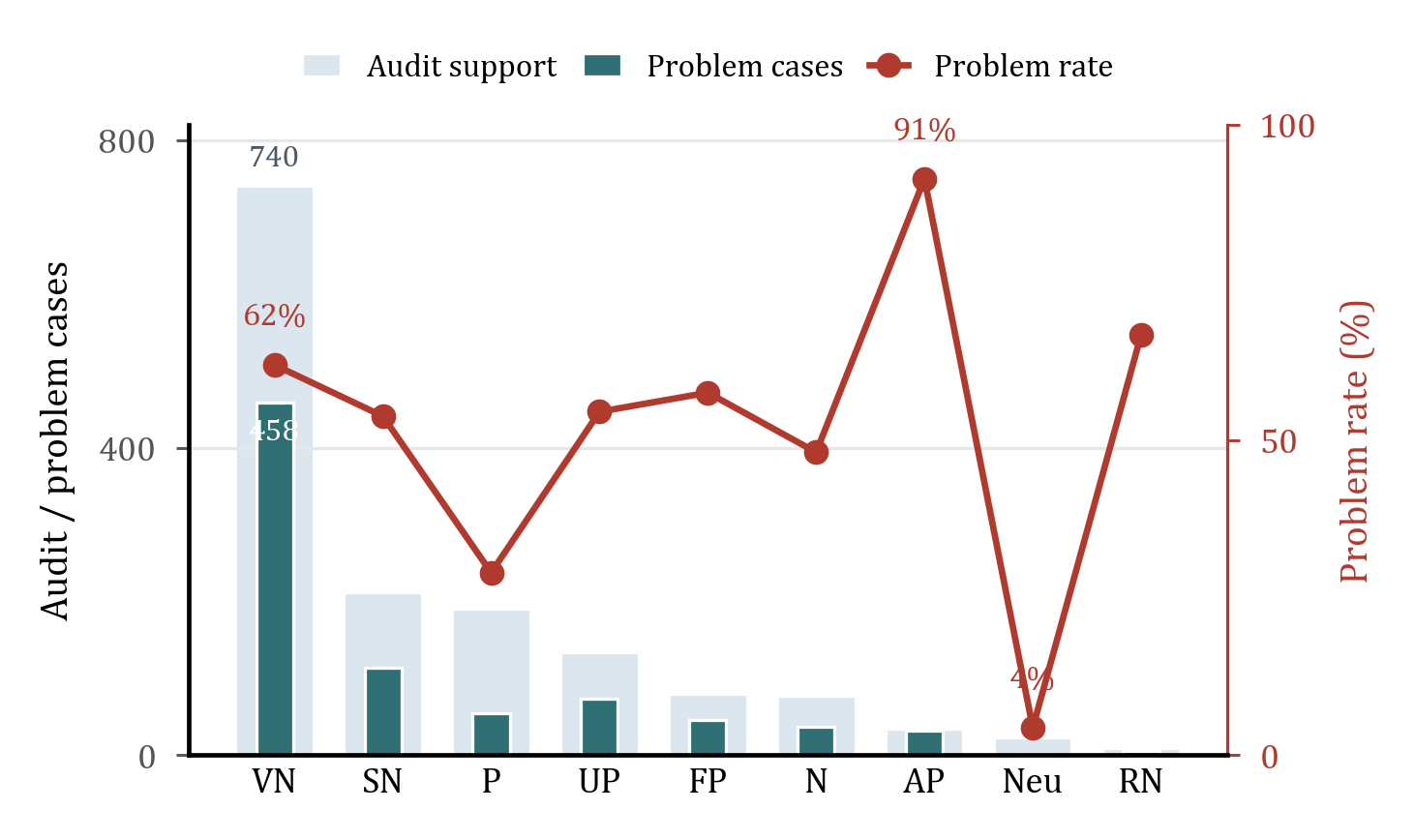}
    \caption{
    Wide light bars show audited support, narrow dark bars show problem-case counts, and the red line shows the problem-case rate.
    }
    \label{fig:annotation_human_ai_friction}
\end{figure}

\paragraph{\textbf{RQ2: Where does human--AI annotation friction arise?}}
We analyze a 1,500-instance audit sample from the model-disagreement pool, where GPT adjudication was run in both candidate orders and compared with human review. 
A case is marked as problematic if the two adjudication orders are inconsistent, or if a consistent GPT decision disagrees with the human-accepted label.

\textbf{Veiled negative cases are the main source of friction.}
As shown in Figure~\ref{fig:annotation_human_ai_friction}, problematic cases concentrate heavily in \textsc{Veiled Negative}: 458 cases, covering 62\% of audited \textsc{Veiled Negative} instances. 
This suggests that neutral-looking utterances with negative implication are especially difficult for AI adjudication.

\textbf{Friction appears when surface affect and implied affect diverge.}
\textsc{Sarcastic Negative} and \textsc{Understated Positive} often require pragmatic reversal or concealment, while \textsc{Formulaic Positive} requires distinguishing genuine positivity from scripted politeness. 
These patterns show that human review is most valuable for stance categories with strong Explicit-Implicit mismatch.


\section{Conclusion}
\label{sec:conclusion}

We introduce CUE-Bench, a Chinese benchmark for unsaid emotion understanding built around the Explicit-Implicit Stance Matrix. By modeling explicit signals, implicit tendencies, Affective Stance, Pragmatic Intent, and Fine-grained Emotion, CUE-Bench enables structured evaluation of affective meaning beyond surface expressions. Experiments show that matrix-guided prompting consistently improves performance, while oracle-conditioning further confirms Affective Stance as a useful intermediate representation for intent and emotion inference. These results show that surface affect recognition alone is insufficient when what speakers mean diverges from what they explicitly say. CUE-Bench thus provides a structured framework for studying and diagnosing polite, suppressed, ironic, indirect, and other forms of unsaid affect in Chinese discourse.





\section{Limitations}
\label{sec:limitations}

CUE-Bench has four main limitations.
\noindent\textbf{(1) Residual annotation noise.}
Although our dual-model annotation pipeline is calibrated and verified with human-labeled data, the final benchmark may still contain unavoidable residual noise. 
In particular, LLM adjudication is used only as a constrained component with consistency filtering, but some uncertain or contaminated instances may remain.
\noindent\textbf{(2) Coarse Explicit-Implicit orientation space.}
The three-way explicit/implicit orientation space makes Affective Stance interpretable and easy to operationalize, but it inevitably abstracts away finer affective distinctions that must be recovered at the Fine-grained Emotion stage.
\noindent\textbf{(3) Long-tailed labels.}
CUE-Bench is long-tailed: categories such as \textsc{Reportive Negative}, \textsc{Empathy}, and several rare emotions have limited support, so macro-F1 and weighted-F1 should be read together. 
Moreover, implicit affect and pragmatic intent remain partly subjective, even with detailed guidelines and adjudication.
\noindent\textbf{(4) Linguistic and cultural scope.}
CUE-Bench currently focuses on Chinese discourse.
Although similar surface--implicit affect relations exist across languages and cultures, multilingual extensions require localized label definitions, annotation guidelines, and re-validation, which limits the current benchmark's generality.




\section*{Acknowledgments}
This work is supported by the National Natural Science Foundation of China (Numbers 62272184 and 62402189),
the China Postdoctoral Science Foundation (Numbers 2024M751012, 2025T180429, and GZC20230894),
the Postdoctor Project of Hubei Province (Number 2024HBBHCXB014),
the Natural Science Foundation of Hubei Province No.JCZRMS202600758,
and CIPS-SMP-Zhipu Large Model Fund (CIPS-SMP20250306).
The computation is completed in the HPC Platform of Huazhong University of Science and Technology.

\bibliography{main}

@inproceedings{wang2020chinese,
  title = {A Large-Scale Chinese Short-Text Conversation Dataset},
  author = {Wang, Yida and Ke, Pei and Zheng, Yinhe and Huang, Kaili and Jiang, Yong and Zhu, Xiaoyan and Huang, Minlie},
  booktitle = {NLPCC},
  year = {2020},
  url = {https://arxiv.org/abs/2008.03946}
}

@inproceedings{chen-etal-2020-jddc,
    title = "The {JDDC} Corpus: A Large-Scale Multi-Turn {C}hinese Dialogue Dataset for {E}-commerce Customer Service",
    author = "Chen, Meng and
      Liu, Ruixue and
      Shen, Lei and
      Yuan, Shaozu and
      Zhou, Jingyan and
      Wu, Youzheng and
      He, Xiaodong and
      Zhou, Bowen",
    booktitle = "Proceedings of the Twelfth Language Resources and Evaluation Conference",
    month = may,
    year = "2020",
    address = "Marseille, France",
    publisher = "European Language Resources Association",
    url = "https://aclanthology.org/2020.lrec-1.58/",
    pages = "459--466"
}

@misc{ccac2024-chinese-sarcasm-calculation,
  title = {Chinese Sarcasm Calculation Evaluation Task at {CCAC} 2024},
  author = {{CCAC 2024 Chinese Sarcasm Calculation Organizers}},
  year = {2024},
  howpublished = {\url{https://github.com/pjzj220113/chinese-sarcasm-calculation}},
  note = {Evaluation task dataset and instructions}
}

@misc{devon018-cn-sarcasmbench,
  title = {{CN-SarcasmBench}},
  author = {{Devon018}},
  year = {2025},
  howpublished = {\url{https://huggingface.co/datasets/Devon018/CN-SarcasmBench}},
  note = {Hugging Face dataset; licensed under CC-BY-NC-4.0}
}

@inproceedings{wang-etal-2021-cnewsum,
  author = {Wang, Danqing and Chen, Jiaze and Wu, Xianze and Zhou, Hao and Li, Lei},
  title = {{CNewSum}: A Large-Scale Summarization Dataset with Human-Annotated Adequacy and Deducibility Level},
  booktitle = {Natural Language Processing and Chinese Computing},
  year = {2021},
  publisher = {Springer International Publishing},
  address = {Cham},
  pages = {389--400},
  doi = {10.1007/978-3-030-88480-2_31}
}

@article{zhang-etal-2016-cec-corpus,
  title = {Event Recognition Based on Deep Learning in Chinese Texts},
  author = {Zhang, Yajun and Liu, Zongtian and Zhou, Wen},
  journal = {PLOS ONE},
  volume = {11},
  number = {8},
  pages = {e0160147},
  year = {2016},
  doi = {10.1371/journal.pone.0160147},
  note = {Data available at \url{https://github.com/shijiebei2009/CEC-Corpus}}
}

@inproceedings{dailydialog-Multi-turn-Dialogue-Dataset-AFNLP2017,
    title = "{D}aily{D}ialog: A Manually Labelled Multi-turn Dialogue Dataset",
    author = "Li, Yanran  and
      Su, Hui  and
      Shen, Xiaoyu  and
      Li, Wenjie  and
      Cao, Ziqiang  and
      Niu, Shuzi",
    editor = "Kondrak, Greg  and
      Watanabe, Taro",
    booktitle = "Proceedings of the Eighth International Joint Conference on Natural Language Processing (Volume 1: Long Papers)",
    month = nov,
    year = "2017",
    address = "Taipei, Taiwan",
    publisher = "Asian Federation of Natural Language Processing",
    url = "https://aclanthology.org/I17-1099/",
    pages = "986--995"
}

@inproceedings{goemotions-Fine-Grained-Emotions-acl2020,
    title = "{G}o{E}motions: A Dataset of Fine-Grained Emotions",
    author = "Demszky, Dorottya  and
      Movshovitz-Attias, Dana  and
      Ko, Jeongwoo  and
      Cowen, Alan  and
      Nemade, Gaurav  and
      Ravi, Sujith",
    editor = "Jurafsky, Dan  and
      Chai, Joyce  and
      Schluter, Natalie  and
      Tetreault, Joel",
    booktitle = "Proceedings of the 58th Annual Meeting of the Association for Computational Linguistics",
    month = jul,
    year = "2020",
    address = "Online",
    publisher = "Association for Computational Linguistics",
    url = "https://aclanthology.org/2020.acl-main.372/",
    doi = "10.18653/v1/2020.acl-main.372",
    pages = "4040--4054"
}

@inproceedings{iest-Implicit-Emotions-acl2018,
    title = "{IEST}: {WASSA}-2018 Implicit Emotions Shared Task",
    author = "Klinger, Roman  and
      De Clercq, Orph{\'e}e  and
      Mohammad, Saif  and
      Balahur, Alexandra",
    editor = "Balahur, Alexandra  and
      Mohammad, Saif M.  and
      Hoste, Veronique  and
      Klinger, Roman",
    booktitle = "Proceedings of the 9th Workshop on Computational Approaches to Subjectivity, Sentiment and Social Media Analysis",
    month = oct,
    year = "2018",
    address = "Brussels, Belgium",
    publisher = "Association for Computational Linguistics",
    url = "https://aclanthology.org/W18-6206/",
    doi = "10.18653/v1/W18-6206",
    pages = "31--42"
}

@inproceedings{resemo-socialMediaContent-emotion-aclfinding2024,
    title = "{RESEMO}: A Benchmark {C}hinese Dataset for Studying Responsive Emotion from Social Media Content",
    author = "Hu, Bo  and
      Zhang, Meng  and
      Xie, Chenfei  and
      Tian, Yuanhe  and
      Song, Yan  and
      Mao, Zhendong",
    editor = "Ku, Lun-Wei  and
      Martins, Andre  and
      Srikumar, Vivek",
    booktitle = "Findings of the Association for Computational Linguistics: ACL 2024",
    month = aug,
    year = "2024",
    address = "Bangkok, Thailand",
    publisher = "Association for Computational Linguistics",
    url = "https://aclanthology.org/2024.findings-acl.970/",
    doi = "10.18653/v1/2024.findings-acl.970",
    pages = "16375--16387"
}

@inproceedings{SMP2020-EWECT,
    title = "Emotion Classification of {COVID}-19 {C}hinese Microblogs based on the Emotion Category Description",
    author = "Xianwei, Guo  and
      Hua, Lai  and
      Yan, Xiang  and
      Zhengtao, Yu  and
      Yuxin, Huang",
    editor = "Li, Sheng  and
      Sun, Maosong  and
      Liu, Yang  and
      Wu, Hua  and
      Liu, Kang  and
      Che, Wanxiang  and
      He, Shizhu  and
      Rao, Gaoqi",
    booktitle = "Proceedings of the 20th Chinese National Conference on Computational Linguistics",
    month = aug,
    year = "2021",
    address = "Huhhot, China",
    publisher = "Chinese Information Processing Society of China",
    url = "https://aclanthology.org/2021.ccl-1.82/",
    pages = "916--927",
    language = "eng"
}

@article{plutchik2001nature,
  title={The nature of emotions: Human emotions have deep evolutionary roots, a fact that may explain their complexity and provide tools for clinical practice},
  author={Plutchik, Robert},
  journal={American scientist},
  volume={89},
  number={4},
  pages={344--350},
  year={2001},
  publisher={JSTOR}
}

@inproceedings{MELD-Multimodal-Emotional-Dialogue-ACL2019,
    title = "{MELD}: A Multimodal Multi-Party Dataset for Emotion Recognition in Conversations",
    author = "Poria, Soujanya  and
      Hazarika, Devamanyu  and
      Majumder, Navonil  and
      Naik, Gautam  and
      Cambria, Erik  and
      Mihalcea, Rada",
    editor = "Korhonen, Anna  and
      Traum, David  and
      M{\`a}rquez, Llu{\'i}s",
    booktitle = "Proceedings of the 57th Annual Meeting of the Association for Computational Linguistics",
    month = jul,
    year = "2019",
    address = "Florence, Italy",
    publisher = "Association for Computational Linguistics",
    url = "https://aclanthology.org/P19-1050/",
    doi = "10.18653/v1/P19-1050",
    pages = "527--536"
}

@misc{CPED-PersonalizedEmotionalDialogue,
      title={CPED: A Large-Scale Chinese Personalized and Emotional Dialogue Dataset for Conversational AI}, 
      author={Yirong Chen and Weiquan Fan and Xiaofen Xing and Jianxin Pang and Minlie Huang and Wenjing Han and Qianfeng Tie and Xiangmin Xu},
      year={2022},
      eprint={2205.14727},
      archivePrefix={arXiv},
      primaryClass={cs.CL},
      url={https://arxiv.org/abs/2205.14727}, 
}

@article{CMMA-multimodal-multiAffection-Nips2023,
  title={CMMA: benchmarking multi-affection detection in chinese multi-modal conversations},
  author={Zhang, Yazhou and Yu, Yang and Guo, Qing and Wang, Benyou and Zhao, Dongming and Uprety, Sagar and Song, Dawei and Li, Qiuchi and Qin, Jing},
  journal={Advances in Neural Information Processing Systems},
  volume={36},
  pages={18794--18805},
  year={2023}
}

@article{du2007stance,
  title={The stance triangle},
  author={Du Bois, John W},
  journal={Stancetaking in discourse: Subjectivity, evaluation, interaction},
  volume={164},
  number={3},
  pages={139--182},
  year={2007}
}

@article{Diplomat-Nips2023,
  title={Diplomat: A dialogue dataset for situated pragmatic reasoning},
  author={Li, Hengli and Zhu, Song-Chun and Zheng, Zilong},
  journal={Advances in Neural Information Processing Systems},
  volume={36},
  pages={46856--46884},
  year={2023}
}

@inproceedings{PUB-PragmaticsUnderstanding-aclfinding2024,
    title = "{PUB}: A Pragmatics Understanding Benchmark for Assessing {LLM}s' Pragmatics Capabilities",
    author = "Sravanthi, Settaluri  and
      Doshi, Meet  and
      Tankala, Pavan  and
      Murthy, Rudra  and
      Dabre, Raj  and
      Bhattacharyya, Pushpak",
    editor = "Ku, Lun-Wei  and
      Martins, Andre  and
      Srikumar, Vivek",
    booktitle = "Findings of the Association for Computational Linguistics: ACL 2024",
    month = aug,
    year = "2024",
    address = "Bangkok, Thailand",
    publisher = "Association for Computational Linguistics",
    url = "https://aclanthology.org/2024.findings-acl.719/",
    doi = "10.18653/v1/2024.findings-acl.719",
    pages = "12075--12097"
}

@article{text-annotation,
   title={ChatGPT outperforms crowd workers for text-annotation tasks},
   volume={120},
   ISSN={1091-6490},
   url={http://dx.doi.org/10.1073/pnas.2305016120},
   DOI={10.1073/pnas.2305016120},
   number={30},
   journal={Proceedings of the National Academy of Sciences},
   publisher={Proceedings of the National Academy of Sciences},
   author={Gilardi, Fabrizio and Alizadeh, Meysam and Kubli, Maël},
   year={2023},
   month=Jul }

@inproceedings{LLMasJudge-nips2023,
    title = {Judging LLM-as-a-judge with MT-bench and Chatbot Arena},
    author = {Zheng, Lianmin and Chiang, Wei-Lin and Sheng, Ying and Zhuang, Siyuan and Wu, Zhanghao and Zhuang, Yonghao and Lin, Zi and Li, Zhuohan and Li, Dacheng and Xing, Eric P. and Zhang, Hao and Gonzalez, Joseph E. and Stoica, Ion},
    year = {2023},
    publisher = {Curran Associates Inc.},
    address = {Red Hook, NY, USA},
    booktitle = {Proceedings of the 37th International Conference on Neural Information Processing Systems},
    articleno = {2020},
    numpages = {29},
    location = {New Orleans, LA, USA},
    series = {NIPS '23}
}

@inproceedings{Cross-replication-Reliability-acl2021,
    title = "Cross-replication Reliability - An Empirical Approach to Interpreting Inter-rater Reliability",
    author = "Wong, Ka  and
      Paritosh, Praveen  and
      Aroyo, Lora",
    editor = "Zong, Chengqing  and
      Xia, Fei  and
      Li, Wenjie  and
      Navigli, Roberto",
    booktitle = "Proceedings of the 59th Annual Meeting of the Association for Computational Linguistics and the 11th International Joint Conference on Natural Language Processing (Volume 1: Long Papers)",
    month = aug,
    year = "2021",
    address = "Online",
    publisher = "Association for Computational Linguistics",
    url = "https://aclanthology.org/2021.acl-long.548/",
    doi = "10.18653/v1/2021.acl-long.548",
    pages = "7053--7065"
}

@inproceedings{du-hoste-2025-another,
    title = "Another Approach to Agreement Measurement and Prediction with Emotion Annotations",
    author = "Du, Quanqi  and
      Hoste, Veronique",
    editor = "Peng, Siyao  and
      Rehbein, Ines",
    booktitle = "Proceedings of the 19th Linguistic Annotation Workshop (LAW-XIX-2025)",
    month = jul,
    year = "2025",
    address = "Vienna, Austria",
    publisher = "Association for Computational Linguistics",
    url = "https://aclanthology.org/2025.law-1.7/",
    doi = "10.18653/v1/2025.law-1.7",
    pages = "87--102",
    ISBN = "979-8-89176-262-6"
}

@inproceedings{Bert-naacl2019,
    title = "{BERT}: Pre-training of Deep Bidirectional Transformers for Language Understanding",
    author = "Devlin, Jacob  and
      Chang, Ming-Wei  and
      Lee, Kenton  and
      Toutanova, Kristina",
    editor = "Burstein, Jill  and
      Doran, Christy  and
      Solorio, Thamar",
    booktitle = "Proceedings of the 2019 Conference of the North {A}merican Chapter of the Association for Computational Linguistics: Human Language Technologies, Volume 1 (Long and Short Papers)",
    month = jun,
    year = "2019",
    address = "Minneapolis, Minnesota",
    publisher = "Association for Computational Linguistics",
    url = "https://aclanthology.org/N19-1423/",
    doi = "10.18653/v1/N19-1423",
    pages = "4171--4186"
}

@article{maskBert,
  title={Pre-training with whole word masking for chinese bert},
  author={Cui, Yiming and Che, Wanxiang and Liu, Ting and Qin, Bing and Yang, Ziqing},
  journal={IEEE/ACM transactions on audio, speech, and language processing},
  volume={29},
  pages={3504--3514},
  year={2021},
  publisher={IEEE}
}

@book{picard1997affective,
  title={Affective Computing},
  author={Picard, Rosalind W},
  year={1997},
  publisher={The MIT Press}
}

@book{brown1987politeness,
  title={Politeness: Some universals in language usage},
  author={Brown, Penelope and Levinson, Stephen C},
  volume={4},
  year={1987},
  publisher={Cambridge university press}
}

@incollection{grice1975logic,
  title={Logic and conversation},
  author={Grice, Herbert P},
  booktitle={Speech acts},
  pages={41--58},
  year={1975},
  publisher={Brill}
}

@article{gross1998emerging,
  title={The emerging field of emotion regulation: An integrative review},
  author={Gross, James J},
  journal={Review of general psychology},
  volume={2},
  number={3},
  pages={271--299},
  year={1998},
  publisher={SAGE Publications Sage CA: Los Angeles, CA}
}

@inproceedings{NTUA-SLP-IEST2018,
    title = "{NTUA}-{SLP} at {IEST} 2018: Ensemble of Neural Transfer Methods for Implicit Emotion Classification",
    author = "Chronopoulou, Alexandra  and
      Margatina, Aikaterini  and
      Baziotis, Christos  and
      Potamianos, Alexandros",
    editor = "Balahur, Alexandra  and
      Mohammad, Saif M.  and
      Hoste, Veronique  and
      Klinger, Roman",
    booktitle = "Proceedings of the 9th Workshop on Computational Approaches to Subjectivity, Sentiment and Social Media Analysis",
    month = oct,
    year = "2018",
    address = "Brussels, Belgium",
    publisher = "Association for Computational Linguistics",
    url = "https://aclanthology.org/W18-6209/",
    doi = "10.18653/v1/W18-6209",
    pages = "57--64"
}

@article{JointBERT-intent-slot-2019,
    title = "{BERT} for Joint Intent Classification and Slot Filling",
    author = "Chen, Qian and Zhuo, Zhu and Wang, Wen",
    journal = "arXiv preprint arXiv:1902.10909",
    year = "2019",
    url = "https://arxiv.org/abs/1902.10909"
}

@inproceedings{SLIM-multiIntent-2021,
  title={Slim: Explicit slot-intent mapping with bert for joint multi-intent detection and slot filling},
  author={Cai, Fengyu and Zhou, Wanhao and Mi, Fei and Faltings, Boi},
  booktitle={ICASSP 2022-2022 IEEE International Conference on Acoustics, Speech and Signal Processing (ICASSP)},
  pages={7607--7611},
  year={2022},
  organization={IEEE}
}

@inproceedings{COSMIC-emnlp2020,
    title = "{COSMIC}: {CO}mmon{S}ense knowledge for e{M}otion Identification in Conversations",
    author = "Ghosal, Deepanway  and
      Majumder, Navonil  and
      Gelbukh, Alexander  and
      Mihalcea, Rada  and
      Poria, Soujanya",
    editor = "Cohn, Trevor  and
      He, Yulan  and
      Liu, Yang",
    booktitle = "Findings of the Association for Computational Linguistics: EMNLP 2020",
    month = nov,
    year = "2020",
    address = "Online",
    publisher = "Association for Computational Linguistics",
    url = "https://aclanthology.org/2020.findings-emnlp.224/",
    doi = "10.18653/v1/2020.findings-emnlp.224",
    pages = "2470--2481"
}

@inproceedings{liu2025learning,
  title={Learning to Substitute Words with Model-based Score Ranking},
  author={Liu, Hongye and Henao, Ricardo},
  booktitle={Proceedings of the 2025 Conference of the Nations of the Americas Chapter of the Association for Computational Linguistics: Human Language Technologies (Volume 1: Long Papers)},
  pages={11551--11565},
  year={2025}
}

@article{liu2026learning,
  title={Learning to Control Summaries with Score Ranking},
  author={Liu, Hongye and Ding, Liang and Henao, Ricardo},
  journal={arXiv preprint arXiv:2604.17197},
  year={2026}
}

@inproceedings{logicAgent-2026,
    title = "Semantic-Aware Logical Reasoning via a Semiotic Framework",
    author = "Zhang, Yunyao  and
      Zhang, Xinglang  and
      Sheng, Junxi  and
      Li, Wenbing  and
      Yu, Junqing  and
      Chen, Yi-Ping Phoebe  and
      Yang, Wei  and
      Song, Zikai",
    editor = "Liakata, Maria  and
      Moreira, Viviane P.  and
      Zhang, Jiajun  and
      Jurgens, David",
    booktitle = "Proceedings of the 64th Annual Meeting of the {A}ssociation for {C}omputational {L}inguistics (Volume 1: Long Papers)",
    month = jul,
    year = "2026",
    address = "San Diego, California, United States",
    publisher = "Association for Computational Linguistics",
    url = "https://aclanthology.org/2026.acl-long.835/",
    doi = "10.18653/v1/2026.acl-long.835",
    pages = "18349--18374",
    ISBN = "979-8-89176-390-6"
}

@inproceedings{LPT-2026logical,
    title = "Logical Phase Transitions: Understanding Collapse in {LLM} Logical Reasoning",
    author = "Zhang, Xinglang  and
      Zhang, Yunyao  and
      Chen, ZeLiang  and
      Yu, Junqing  and
      Yang, Wei  and
      Song, Zikai",
    editor = "Liakata, Maria  and
      Moreira, Viviane P.  and
      Zhang, Jiajun  and
      Jurgens, David",
    booktitle = "Proceedings of the 64th Annual Meeting of the {A}ssociation for {C}omputational {L}inguistics (Volume 1: Long Papers)",
    month = jul,
    year = "2026",
    address = "San Diego, California, United States",
    publisher = "Association for Computational Linguistics",
    url = "https://aclanthology.org/2026.acl-long.858/",
    doi = "10.18653/v1/2026.acl-long.858",
    pages = "18836--18860",
    ISBN = "979-8-89176-390-6"
}

@article{ye2026omnitrend,
  title={OmniTrend: Content-Context Modeling for Scalable Social Popularity Prediction},
  author={Ye, Liliang and Zeng, Guiyi and Zhang, Yunyao and Chen, Yi-Ping Phoebe and Yu, Junqing and Song, Zikai},
  journal={arXiv preprint arXiv:2604.26252},
  year={2026}
}

@misc{intervenSim-2026,
      title={IntervenSim: Intervention-Aware Social Network Simulation for Opinion Dynamics}, 
      author={Yunyao Zhang and Zuocheng Ying and Xinglang Zhang and Junqing Yu and Peng Fang and Xu Chen and Wei Yang and Zikai Song},
      year={2026},
      eprint={2604.06600},
      archivePrefix={arXiv},
      primaryClass={cs.SI},
      url={https://arxiv.org/abs/2604.06600}, 
}

@misc{ACMMM2026-Popularity,
      title={Seeing Further and Wider: Joint Spatio-Temporal Enlargement for Micro-Video Popularity Prediction}, 
      author={Dali Wang and Yunyao Zhang and Junqing Yu and Yi-Ping Phoebe Chen and Chen Xu and Zikai Song},
      year={2026},
      eprint={2604.20311},
      archivePrefix={arXiv},
      primaryClass={cs.MM},
      url={https://arxiv.org/abs/2604.20311}, 
}

@inproceedings{GAS32025-ga,
    title = "$GA-S^3$: Comprehensive Social Network Simulation with Group Agents",
    author = "Zhang, Yunyao  and
      Song, Zikai  and
      Zhou, Hang  and
      Ren, Wenfeng  and
      Chen, Yi-Ping Phoebe  and
      Yu, Junqing  and
      Yang, Wei",
    editor = "Che, Wanxiang  and
      Nabende, Joyce  and
      Shutova, Ekaterina  and
      Pilehvar, Mohammad Taher",
    booktitle = "Findings of the Association for Computational Linguistics: ACL 2025",
    month = jul,
    year = "2025",
    address = "Vienna, Austria",
    publisher = "Association for Computational Linguistics",
    url = "https://aclanthology.org/2025.findings-acl.468/",
    doi = "10.18653/v1/2025.findings-acl.468",
    pages = "8950--8970",
    ISBN = "979-8-89176-256-5",
}

@article{hotcomment-2026,
  title={HotComment: A Benchmark for Evaluating Popularity of Online Comments},
  author={Wu, Yafeng and Zhang, Yunyao and Ye, Liliang and Zeng, Guiyi and Yu, Junqing and Xu, Chen and Song, Zikai},
  journal={arXiv preprint arXiv:2604.25614},
  year={2026}
}

@misc{song2026socialintelligence,
  title = {Social Intelligence Modeling: A Comprehensive Survey from Social Perception to Social Simulation},
  author = {Song, Zikai and Li, Xiajie and Zhang, Yunyao and Zhang, Xinglang and Yang, Wei and Yu, Junqing},
  howpublished = {ResearchGate preprint},
  year = {2026},
  note = {Preprint available on ResearchGate},
  doi = {10.13140/RG.2.2.21157.87528},
  url = {https://doi.org/10.13140/RG.2.2.21157.87528}
}

@inproceedings{ICLR2026_LoRA-Mixer,
 author = {Li, Wenbing and Song, Zikai and Zhou, Hang and Yu, Junqing and Zhang, Yunyao and Yang, Wei},
 booktitle = {International Conference on Learning Representations},
 editor = {C. Vondrick and B. Hariharan and C. Raffel and L. Pinto and D. Yang and A. Faust},
 pages = {14694--14716},
 title = {LoRA-Mixer: Coordinate Modular LoRA Experts Through Serial Attention Routing},
 url = {https://proceedings.iclr.cc/paper_files/paper/2026/file/18610cbcd1da57854aa05ebdc5cd3168-Paper-Conference.pdf},
 volume = {2026},
 year = {2026}
}

@article{1COMBINER,
  title={COMBINER: Composed Image Retrieval Guided by Attribute-based Neighbor Relations},
  author={Li, Zixu and Hu, Yupeng and Chen, Zhiwei and Wen, Haokun and Song, Xuemeng and Nie, Liqiang},
  journal={IEEE TIP},
  year={2026},
  publisher={IEEE}
}

@inproceedings{4TEMA,
  title={Tema: Anchor the image, follow the text for multi-modification composed image retrieval},
  author={Li, Zixu and Hu, Yupeng and Fu, Zhiheng and Chen, Zhiwei and Li, Yongqi and Nie, Liqiang},
  booktitle={Proceedings of the 64th Annual Meeting of the Association for Computational Linguistics (Volume 1: Long Papers)},
  pages={24421--24442},
  year={2026}
}

@inproceedings{6OFFSET, 
  title = {OFFSET: Segmentation-based Focus Shift Revision for Composed Image Retrieval}, 
  author = {Chen, Zhiwei and Hu, Yupeng and Li, Zixu and Fu, Zhiheng and Song, Xuemeng and Nie, Liqiang}, 
  booktitle = {ACM MM}, 
  pages = {6113–6122}, 
  year = {2025}
}

@article{7REFINE,
  title={REFINE: Composed Video Retrieval via Shared and Differential Semantics Enhancement},
  author={Hu, Yupeng and Li, Zixu and Chen, Zhiwei and Huang, Qinlei and Fu, Zhiheng and Xu, Mingzhu and Nie, Liqiang},
  journal={ACM ToMM},
  year={2026},
  publisher={ACM New York, NY}
}

@inproceedings{8INTENT,
  title={INTENT: Invariance and Discrimination-aware Noise Mitigation for Robust Composed Image Retrieval},
  author={Chen, Zhiwei and Hu, Yupeng and Fu, Zhiheng and Li, Zixu and Huang, Jiale and Huang, Qinlei and Wei, Yinwei},
  booktitle={AAAI},
  volume={40},
  pages={20463--20471},
  year={2026}
}

@article{10STABLE,
  title={STABLE: Efficient Hybrid Nearest Neighbor Search via Magnitude-Uniformity and Cardinality-Robustness},
  author={Yang, Qianyun and Chen, Zhiwei and Hu, Yupeng and Li, Zixu and Fu, Zhiheng and Nie, Liqiang},
  journal={IEEE TKDE},
  year={2026},
  publisher={IEEE}
}

@article{11EgoAction,
  title={EgoAction: Egocentric Action Composition with Reliability-Aware Temporal Fusion for the EPIC-KITCHENS Action Detection Challenge at CVPR 2026},
  author={Fu, Zhiheng and Li, Zixu and Chen, Zhiwei and Liu, Fangxu and Hu, Yupeng and Guan, Weili and Nie, Liqiang},
  journal={arXiv preprint arXiv:2605.24496},
  year={2026}
}

@article{15R3,
  title={R$^3$: Composed Video Retrieval via Reasoning-Guided Recalling and Re-ranking}, 
  author={Zixu Li and Yupeng Hu and Zhiheng Fu and Zhiwei Chen and Weili Guan and Liqiang Nie},
  journal={arXiv preprint arXiv:2606.01113},
  year={2026}
}

@inproceedings{dai-etal-2026-tears,
  title = "Tears or Cheers? Benchmarking {LLM}s via Culturally Elicited Distinct Affective Responses",
  author = "Dai, Chongyuan and Shen, Yaling and Gao, Zihan and Li, Jia and Jiang, Yishun and Wang, Yaxiong and Liu, Liu and Ge, Zongyuan and Hu, Jinpeng",
  editor = "Liakata, Maria and Moreira, Viviane P. and Zhang, Jiajun and Jurgens, David",
  booktitle = "Proceedings of the 64th Annual Meeting of the {A}ssociation for {C}omputational {L}inguistics (Volume 1: Long Papers)",
  month = jul,
  year = "2026",
  address = "San Diego, California, United States",
  publisher = "Association for Computational Linguistics",
  url = "https://aclanthology.org/2026.acl-long.1769/",
  doi = "10.18653/v1/2026.acl-long.1769",
  pages = "38171--38196",
  ISBN = "979-8-89176-390-6"
}

@inproceedings{dai-etal-2026-psyche,
  title = "Psyche-R1: Towards Reliable Psychological {LLM}s through Unified Empathy, Expertise, and Reasoning",
  author = "Dai, Chongyuan and Hu, Jinpeng and Shi, Hongchang and Li, Zhuo and Guo, Dan and Yang, Xun and Wang, Meng",
  editor = "Liakata, Maria and Moreira, Viviane P. and Zhang, Jiajun and Jurgens, David",
  booktitle = "Proceedings of the 64th Annual Meeting of the {A}ssociation for {C}omputational {L}inguistics (Volume 1: Long Papers)",
  month = jul,
  year = "2026",
  address = "San Diego, California, United States",
  publisher = "Association for Computational Linguistics",
  url = "https://aclanthology.org/2026.acl-long.1141/",
  doi = "10.18653/v1/2026.acl-long.1141",
  pages = "24889--24906",
  ISBN = "979-8-89176-390-6"
}

@misc{dai2026dontboxin,
  title = "Don't Box Me In: Dynamic Cultural Adaptation and Cognitive Tracking for Social Understanding",
  author = "Dai, Chongyuan and Shen, Yaling and Tang, Shengeng and Ma, Hui and Hu, Jinpeng",
  year = "2026",
  eprint = "2608.22411",
  archivePrefix = "arXiv",
  primaryClass = "cs.CL",
  url = "https://arxiv.org/abs/2608.22411"
}

@inproceedings{dong-etal-2026-mitigating,
  title = "Mitigating Safety Context Amnesia in Multimodal Reasoning Models via Intent-Guided Safety Reasoning",
  author = "Dong, Xiyao and Cheng, Guangsheng and Chen, YiLong and Zhang, Xiaojin and He, Kun",
  editor = "Liakata, Maria and Moreira, Viviane P. and Zhang, Jiajun and Jurgens, David",
  booktitle = "Proceedings of the 64th Annual Meeting of the {A}ssociation for {C}omputational {L}inguistics (Volume 1: Long Papers)",
  month = jul,
  year = "2026",
  address = "San Diego, California, United States",
  publisher = "Association for Computational Linguistics",
  url = "https://aclanthology.org/2026.acl-long.1821/",
  doi = "10.18653/v1/2026.acl-long.1821",
  pages = "39249--39276",
  ISBN = "979-8-89176-390-6"
}

@article{liao2026unlocking,
  title={Unlocking explainable and effective multimodal affective reasoning via large language models},
  author={Liao, Junjie and Zeng, Jiandian and Song, Binbin and Zhou, Mengting and Fan, Xiaopeng and Wang, Tian},
  journal={Pattern Recognition},
  pages={113366},
  year={2026},
  publisher={Elsevier}
}

@inproceedings{liao2026role,
  title={How do Role Models Shape Collective Morality? Exemplar-Driven Moral Learning in Multi-Agent Simulation},
  author={Liao, Junjie and Tang, Huacong and Ziheng, Zhou and Wang, Yizhou and Zhong, Fangwei},
  booktitle={Proceedings of the 64th Annual Meeting of the Association for Computational Linguistics (Volume 1: Long Papers)},
  pages={42981--43016},
  year={2026}
}

\appendix
\clearpage
\section*{\centering Appendix}

This appendix provides implementation details for annotation, examples, prompts, and release documentation.

\section*{The Usage of LLM}
In accordance with ACL policy, LLMs were used as writing and annotation-support tools. For dataset construction, model outputs are treated as candidate labels and rationales. Final labels are human-validated. 

\section{Additional Distributional Analysis}
\label{app:distribution_analysis}

\begin{figure*}[t] 
   \centering 
   \includegraphics[width=0.98\textwidth]{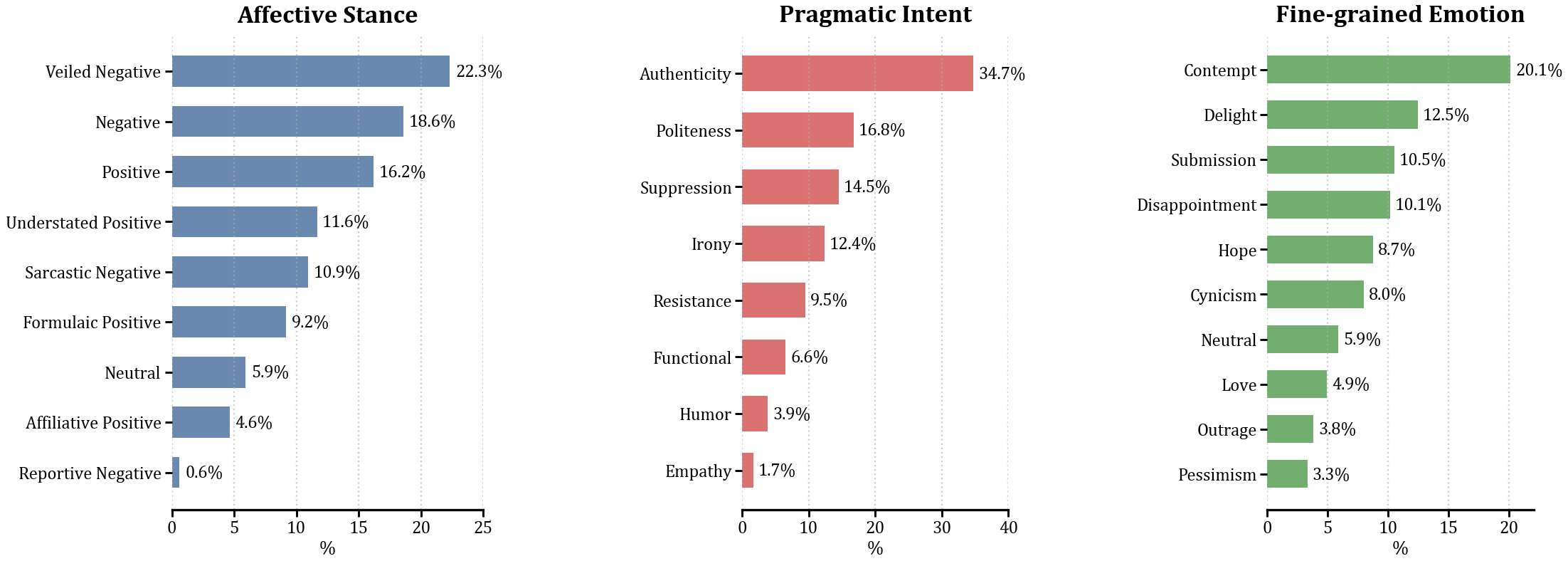} 
   \caption{ 
   Distribution of the three label layers in CUE-Bench. Affective Stance and Pragmatic Intent are shown in full; Fine-grained Emotion shows the ten most frequent categories for readability. 
   } 
   \label{fig:label_distribution} 
\end{figure*}

\paragraph{Context-bound stance categories.}
The rarest Affective Stance is \textsc{Reportive Negative} (0.6\%), where negative surface wording is used with a largely neutral, reportive force.
This pattern is more natural in news-style reporting, incident summaries, or analytical long-form writing than in casual online interaction.
Because CUE-Bench is dominated by internet dialogue and social-media discourse~\cite{hotcomment-2026}, \textsc{Reportive Negative} remains sparse; Zhihu-style long-form text is more likely to contain such cases, but it is not the dominant source.
By contrast, \textsc{Formulaic Positive} (9.2\%) is tied to service-oriented interaction, where thanks, apologies, honorifics, and blessing formulas are often routine rather than deeply positive.

\paragraph{Intent and emotion distributions.}
\textsc{Authenticity} is the most frequent Pragmatic Intent (34.7\%), while socially mediated intents such as \textsc{Politeness} (16.8\%), \textsc{Suppression} (14.5\%), \textsc{Irony} (12.4\%), and \textsc{Resistance} (9.5\%) remain substantial.
Fine-grained Emotion shows a stronger long tail.
The right panel of Figure~\ref{fig:label_distribution} shows the high-frequency head; low-frequency emotions are omitted from the visualization for readability but remain part of the benchmark and evaluation.
The emotion inventory is organized with reference to Plutchik's emotion wheel~\citep{plutchik2001nature}.
The prominence of \textsc{Contempt} (20.1\%) is consistent with the negative and sarcastic bias of the initial corpus construction, while categories such as \textsc{Submission} and \textsc{Hope} show that the benchmark also captures relational and anticipatory affect.

\section{Additional Ablation: Structure-only Baseline}
\label{app:structure_only}

To test whether the gains come merely from using a structured prompt, we add a Structure-only baseline on DeepSeek-V4-Flash.
This baseline keeps a generic step-by-step reasoning format for downstream prediction, but does not use our Explicit-Implicit Stance Matrix.
As shown in Table~\ref{tab:structure_only_ablation}, Structure-only remains close to free-form CoT, while Matrix-guided prompting is substantially stronger on Pragmatic Intent and Fine-grained Emotion.
This indicates that the gains mainly come from the stance matrix rather than from the generic step-by-step format itself.

\begin{table}[H]
\centering
\footnotesize
\setlength{\tabcolsep}{4pt}
\renewcommand{\arraystretch}{1.10}
\begin{tabular}{llccc}
\toprule
\textbf{Task} & \textbf{Method} & \textbf{Acc.} & \textbf{Ma-F1} & \textbf{W-F1} \\
\midrule
\multirow{4}{*}{Intent}
& CoT & 0.332 & 0.277 & 0.343 \\
& Structure-only & 0.333 & 0.274 & 0.341 \\
& \cellcolor{gray!12}Matrix-guided
& \cellcolor{gray!12}\textbf{0.459}
& \cellcolor{gray!12}\textbf{0.361}
& \cellcolor{gray!12}\textbf{0.465} \\
& $\Delta$ & \posdelta{+0.126} & \posdelta{+0.087} & \posdelta{+0.124} \\
\midrule
\multirow{4}{*}{Emotion}
& CoT & 0.251 & 0.182 & 0.253 \\
& Structure-only & 0.252 & 0.170 & 0.239 \\
& \cellcolor{gray!12}Matrix-guided
& \cellcolor{gray!12}\textbf{0.314}
& \cellcolor{gray!12}\textbf{0.194}
& \cellcolor{gray!12}\textbf{0.327} \\
& $\Delta$ & \posdelta{+0.062} & \posdelta{+0.024} & \posdelta{+0.088} \\
\bottomrule
\end{tabular}
\caption{Structure-only ablation on DeepSeek-V4-Flash. Delta rows are computed over Structure-only.}
\label{tab:structure_only_ablation}
\end{table}

\section{CUE-Bench Details}
\label{app:dataset_processing}

\subsection{Release Format and Data Split}

The released data follow a unified format that supports all three benchmark tasks.
Table~\ref{tab:data_fields} summarizes the core fields.

We split the dataset by source instance rather than by individual text span to avoid context leakage across training, development, and test sets.
The split is stratified by Affective Stance where possible, since the nine-way stance distribution is naturally imbalanced.

\subsection{Source Data and Label Independence}

CUE-Bench draws raw Chinese text and dialogue context from LCCC~\citep{wang2020chinese}, JDDC~\citep{chen-etal-2020-jddc}, the CCAC 2024 Chinese Sarcasm Calculation dataset~\citep{ccac2024-chinese-sarcasm-calculation}, Zhihu QA\footnote{\url{https://huggingface.co/datasets/zirui3/zhihu_qa}}, and CN-SarcasmBench~\citep{devon018-cn-sarcasmbench}.
The refined v3.1 release additionally incorporates news-style candidates from CNewSum~\citep{wang-etal-2021-cnewsum} and CEC-Corpus~\citep{zhang-etal-2016-cec-corpus} to improve coverage of \textsc{Reportive Negative}.
These sources are used only as text pools.
We do not reuse their original sentiment labels, sarcasm labels, dialogue labels, or any other source-provided annotations as CUE-Bench labels.
All released Affective Stance, Pragmatic Intent, and Fine-grained Emotion annotations are produced through our own screening, candidate annotation, human verification, and adjudication pipeline.

\subsection{Annotation Pipeline and Quality Control}
\label{app:annotation_agreement}
We construct CUE-Bench through a hybrid annotation pipeline that combines model-assisted pre-annotation, human verification, and consistency-based LLM adjudication.
Starting from 60,000 candidate instances, DeepSeek-V4-Flash and Qwen2.5-72B-Instruct independently produce candidate annotations.
The 20,000 instances on which the two models agree are retained as high-confidence annotations.
For the remaining 40,000 disagreement cases, we manually verify 10,000 instances to construct the human-verified gold subset.
During this process, annotators select the more appropriate label from the model-generated candidates rather than assuming that all model annotations are incorrect.

To further assess annotation reliability, we sample 300 instances from the human-verified gold subset and ask three expert annotators to independently re-annotate them.
As reported in Table~\ref{tab:iaa_summary}, Affective Stance obtains the strongest agreement, while Pragmatic Intent and Fine-grained Emotion show lower raw agreement due to their greater subjectivity and dependence on latent affective inference.
However, agreement improves substantially when downstream labels are evaluated under consistent Affective Stance.
For Pragmatic Intent, the conditional Krippendorff's $\alpha$ increases to 0.7688.
This supports the role of Affective Stance as an intermediate structure for localizing ambiguity and reducing uncertainty in intent and emotion annotation.

For the remaining 30,000 model-disagreement instances, we use a subsequent adjudication step.
Validation on the human-verified gold subset shows that this adjudication step achieves 89\% accuracy when selecting between two model-generated candidate annotations.
To reduce positional bias, each instance is adjudicated twice with the order of candidate labels reversed.
Only instances with consistent forward--reverse adjudication decisions are retained, yielding 21,823 additional instances and discarding 8,177 uncertain cases.
The final benchmark contains 51,823 instances.

\begin{table}[t]
\centering
\footnotesize
\setlength{\tabcolsep}{3pt}
\renewcommand{\arraystretch}{1.12}
\begin{tabularx}{\linewidth}{@{}
>{\raggedright\arraybackslash\ttfamily}p{0.42\linewidth}
@{\hspace{0.8em}}
>{\raggedright\arraybackslash}X
@{}}
\toprule
\normalfont\textbf{Field} & \textbf{Description} \\
\midrule
sample\_id & Anonymized instance identifier. \\
context & Context used for affective interpretation. \\
target\_text & Target utterance or text span to be labeled. \\
explicit\_layer & Surface affective information. \\
implicit\_layer & Context-inferred implicit affective information. \\
affective\_stance & Nine-way Affective Stance label. \\
pragmatic\_intent & Pragmatic Intent label. \\
fine\_grained\_emotion & Fine-grained Emotion label. \\
\bottomrule
\end{tabularx}
\caption{Core fields in the CUE-Bench release format.}
\label{tab:data_fields}
\end{table}

\begin{craddition}
\subsection{Human Annotation Disagreement}
We further analyze which Affective Stance boundaries account for most human annotator disagreement.
As shown in Table~\ref{tab:stance_boundary_disagreement}, the largest boundary pair is \textsc{Reserved Positive} $\leftrightarrow$ \textsc{Covert Negative}, accounting for 20.9\% of pairwise stance disagreements.
This boundary reflects whether a neutral surface expression implies positive accommodation or veiled negative affect.
Other frequent boundary pairs include \textsc{Aligned Positive} $\leftrightarrow$ \textsc{Detached Positive} (9.1\%), where annotators distinguish sincere positivity from formulaic politeness, and \textsc{Covert Negative} $\leftrightarrow$ \textsc{Aligned Negative} (9.1\%), where annotators distinguish veiled negativity from direct negativity.
These disagreements are concentrated around interpretable boundary cases rather than being randomly distributed across the label space.
Such high-disagreement boundary cases can be used for annotator calibration and for further refining label definitions.

\begin{table}[t]
\centering
\footnotesize
\setlength{\tabcolsep}{4pt}
\renewcommand{\arraystretch}{1.1}
\begin{tabularx}{\linewidth}{@{}p{0.36\linewidth}cX@{}}
\toprule
\textbf{Boundary pair} & \textbf{Share} & \textbf{Interpretation} \\
\midrule
\textsc{Reserved Positive} $\leftrightarrow$ \textsc{Covert Negative}
& 20.9\% & Whether neutral wording implies positive accommodation or veiled negative affect. \\
\textsc{Aligned Positive} $\leftrightarrow$ \textsc{Detached Positive}
& 9.1\% & Whether positive wording is sincere or formulaic/polite. \\
\textsc{Covert Negative} $\leftrightarrow$ \textsc{Aligned Negative}
& 9.1\% & Whether negative affect is indirectly implied or directly expressed. \\
\textsc{Aligned Neutral} $\leftrightarrow$ \textsc{Covert Negative}
& 7.5\% & Whether a factual statement carries implicit negative affect. \\
\bottomrule
\end{tabularx}
\caption{Frequent Affective Stance boundary pairs in human annotation disagreement.}
\label{tab:stance_boundary_disagreement}
\end{table}
\end{craddition}

\begin{craddition}
\subsection{Cross-Lingual and Cross-Cultural Transferability}
CUE-Bench focuses on Chinese discourse not because the framework is specific to Chinese, but because Chinese provides rich cases of unsaid meaning, indirect expression, and context-dependent affective communication.
The Explicit-Implicit Stance Matrix abstracts the relation between surface expression and implicit affect, rather than relying on fixed lexical patterns.
Similar relations appear in other languages and cultures, such as sarcasm, understatement, politeness, face-saving, and passive-aggressive expression in English.
However, concrete label definitions, cultural norms, and annotation guidelines should not be transferred word-for-word.
Future multilingual extensions should localize the label inventory and re-validate annotation boundaries for each cultural context while preserving the matrix as an intermediate organizational scaffold.
\end{craddition}

\subsection{Detailed Pairwise Agreement}

Table~\ref{tab:kappa_detailed_appendix} reports pairwise Cohen's $\kappa$ scores between the gold labels and each expert re-annotation.
These results complement the aggregate agreement statistics in Table~\ref{tab:iaa_summary} by showing how each expert annotation compares with the gold labels.
Conditional scores are computed on instances with consistent Affective Stance.

\begin{table}[t]
\centering
\footnotesize
\setlength{\tabcolsep}{4pt}
\renewcommand{\arraystretch}{1.12}
\begin{tabular}{@{}llcc@{}}
\toprule
\textbf{Layer} 
& \textbf{Expert} 
& \textbf{Raw $\kappa$} 
& \textbf{Cond. $\kappa$} \\
\midrule
Affective Stance 
& A & 0.5116 & -- \\
& B & 0.5864 & -- \\
\midrule
Pragmatic Intent 
& A & 0.4057 & 0.8746 \\
& B & 0.3786 & 0.7042 \\
\midrule
Fine-grained Emotion 
& A & 0.3325 & 0.6915 \\
& B & 0.3412 & 0.6462 \\
\bottomrule
\end{tabular}
\caption{
Pairwise Cohen's $\kappa$ scores between the gold labels and expert re-annotations.
Conditional scores are computed on instances with consistent Affective Stance.
}
\label{tab:kappa_detailed_appendix}
\end{table}

\paragraph{Interpreting Agreement Scores.}
We interpret agreement scores in the context of subjective affective annotation rather than relying on a single universal threshold.
Prior work has argued that fixed IRR thresholds such as $\kappa$ or $\alpha > 0.6$ can be overly rigid for subjective tasks with genuine ambiguity~\cite{Cross-replication-Reliability-acl2021}.
Fine-grained emotion annotation also commonly exhibits moderate or low chance-corrected agreement: GoEmotions reports an average Cohen's $\kappa$ of approximately 0.29 across 27 emotion categories~\cite{goemotions-Fine-Grained-Emotions-acl2020}, and recent work on emotion annotation reports Fleiss' $\kappa$ values of 0.19--0.33 and Krippendorff's $\alpha$ values of 0.22--0.64 for emotion and valence annotation settings~\cite{du-hoste-2025-another}.
These findings motivate our use of both raw and conditional agreement scores, where the latter evaluates whether downstream intent and emotion labels become more reliable under a shared Affective Stance interpretation.

\section{Annotation Guidelines}
\label{app:guidelines}

\subsection{Explicit Affective Signal}
Annotators label the explicit affective signal using only the target utterance.
Context may be shown for orientation, but the decision must be justified by surface evidence in the utterance itself.

\noindent\textbf{Positive signal.} Use \textbf{positive} when the utterance directly expresses appreciation, happiness, agreement, praise, relief, or encouragement.

\noindent\textbf{Negative signal.} Use \textbf{negative} when the utterance directly expresses blame, anger, disappointment, sadness, anxiety, refusal, or complaint.

\noindent\textbf{Neutral signal.} Use \textbf{neutral} when the utterance contains no clear surface affective expression, even if the surrounding context is emotional.

\subsection{Implicit Affective Tendency}
Annotators label the implicit affective tendency using the dialogue context and the target utterance together.
The label should capture the affect implied by the speaker's stance, conversational goal, or pragmatic force, rather than simply repeating the surface wording.

\noindent\textbf{Positive tendency.} Use \textbf{positive} when the utterance implies acceptance, care, support, relief, or friendly intent.

\noindent\textbf{Negative tendency.} Use \textbf{negative} when the utterance implies rejection, dissatisfaction, pressure, hostility, disappointment, or sarcasm.

\noindent\textbf{Neutral tendency.} Use \textbf{neutral} when context does not support a clear positive or negative affective inference.

\subsection{Affective Stance Assignment}
Affective Stance is derived from the ordered pair of explicit affective signal and implicit affective tendency.
Annotators therefore do not invent an independent stance label.
They first verify the two base signals, map the pair to the Explicit-Implicit Stance Matrix, and then check whether the resulting stance matches the intended interpretation.
If the mapped stance feels implausible, annotators must revisit the explicit or implicit signal rather than manually overriding the stance.

\subsection{Pragmatic Intent Assignment}
Annotators label Pragmatic Intent as the communicative function or strategy realized by the utterance in context.

\begingroup
\RaggedRight

\noindent\textbf{Authentic expression.} Use \textsc{Authenticity} when the utterance directly expresses the speaker's genuine affective position.

\noindent\textbf{Polite mitigation.} Use \textsc{Politeness} when positive or softened wording primarily serves social etiquette, deference, apology, or service-script politeness.

\noindent\textbf{Affective concealment.} Use \textsc{Suppression} when the speaker hides, weakens, or withholds the implied affect.

\noindent\textbf{Contrastive meaning.} Use \textsc{Irony} when the utterance relies on reversal, sarcasm, exaggerated praise, or contrast between literal and intended meaning.

\noindent\textbf{Oppositional stance.} Use \textsc{Resistance} when the utterance implies refusal, opposition, complaint, pressure, or dissatisfaction.

\noindent\textbf{Task-oriented communication.} Use \textsc{Functional} when the utterance mainly reports, informs, requests, or describes without strong interpersonal strategy.

\noindent\textbf{Playful framing.} Use \textsc{Humor} when the utterance uses playfulness, teasing, or comic framing as the main pragmatic force.

\noindent\textbf{Supportive orientation.} Use \textsc{Empathy} when the utterance primarily conveys care, comfort, solidarity, or perspective-taking toward another person.

\par
\endgroup

\subsection{Fine-grained Emotion Decision Rules}
Annotators label Fine-grained Emotion as the final affective interpretation, using the target utterance, context, Affective Stance, and Pragmatic Intent together.
The label should describe the speaker's inferred affective state, not the emotion that the utterance may cause in the reader.

\noindent\textbf{Specificity preference.} Prefer the most specific emotion supported by contextual evidence; use broad labels such as \textsc{Neutral} only when no specific affect is recoverable.

\noindent\textbf{Underlying affect.} Distinguish outward negativity from the underlying state: complaint may indicate \textsc{Outrage}, \textsc{Disappointment}, \textsc{Contempt}, or \textsc{Anxiety} depending on context.

\noindent\textbf{Relational and future-oriented affect.} Use relational and anticipatory labels such as \textsc{Submission}, \textsc{Hope}, \textsc{Pessimism}, or \textsc{Optimism} when the utterance encodes expectation, dependence, resignation, or future orientation.

\noindent\textbf{Ambiguity resolution.} When multiple emotions are plausible, choose the label best supported by the target utterance and its immediate conversational context, and flag genuinely ambiguous cases for review.
\section{Definitions and Examples of Affective Stances}
\label{app:stance_definitions}

Table~\ref{tab:stance_definitions} provides detailed definitions and examples for the nine Affective Stances in the Explicit-Implicit Stance Matrix.

\begin{CJK*}{UTF8}{gbsn}
\newcolumntype{L}[1]{>{\RaggedRight\arraybackslash}p{#1}}
\newcolumntype{C}[1]{>{\Centering\arraybackslash}p{#1}}
\newcolumntype{Y}{>{\RaggedRight\arraybackslash}X}

\newcommand{\signalpair}[2]{\makecell[c]{$e_i=#1$\\[-1pt]$h_i=#2$}}
\newcommand{\zhen}[2]{\textbf{Zh:} #1\par\textbf{En:} #2}

\begin{table*}[t]
\centering
\small
\setlength{\tabcolsep}{4.5pt}
\renewcommand{\arraystretch}{1.18}

\begin{tabularx}{\textwidth}{@{}c c L{0.15\textwidth} L{0.30\textwidth} X@{}}
\toprule
\textbf{$e_i$} & \textbf{$h_i$}
& \textbf{Affective Stance}
& \textbf{Typical Phenomenon}
& \textbf{Chinese--English Example} \\
\midrule

$+$ & $+$
& \textsc{Positive}
& Direct praise, joy, gratitude, approval, support, or affection.
& \textbf{Zh:} 太好了！你终于熬出来了，我真的为你骄傲。 \newline
  \textbf{En:} That's wonderful! You finally made it through. I'm truly proud of you. \\

\addlinespace[2pt]

$+$ & $0$
& \textsc{Formulaic \newline Positive}
& Customer-service politeness, routine thanks, greetings, scripted apologies, or socially expected positive wording.
& \textbf{Zh:} 亲，真的非常抱歉给您带来不便，感谢您的理解，祝您生活愉快。 \newline
  \textbf{En:} Dear customer, we sincerely apologize for the inconvenience. Thank you for your understanding, and have a nice day. \\

\addlinespace[2pt]

$+$ & $-$
& \textsc{Sarcastic \newline Negative}
& Sarcasm, irony, backhanded praise, mock compliment, or exaggerated praise used to express criticism.
& \textbf{Zh:} 你可真是天才，每次都能精准踩雷。 \newline
  \textbf{En:} What a genius you are, managing to step on the exact landmine every time. \\

\addlinespace[2pt]

$0$ & $+$
& \textsc{Understated \newline Positive}
& Modesty, humblebragging, restrained pride, indirect approval, or neutral wording that hides satisfaction.
& \textbf{Zh:} 也就一般吧，省赛第一而已。 \newline
  \textbf{En:} It was nothing special, just first place in the provincial contest. \\

\addlinespace[2pt]

$0$ & $0$
& \textsc{Neutral}
& Factual, procedural, descriptive, or informational utterances without clear affective commitment.
& \textbf{Zh:} 会议改到下午三点，地点不变。 \newline
  \textbf{En:} The meeting has been moved to 3 p.m.; the location remains unchanged. \\

\addlinespace[2pt]

$0$ & $-$
& \textsc{Veiled \newline Negative}
& Neutral wording that implies dissatisfaction, reluctance, compromise, pressure, rejection, or concealed criticism.
& \textbf{Zh:} 行，你继续按这个方案来，结果我就不评价了。 \newline
  \textbf{En:} Fine, keep following this plan. I won't comment on the result. \\

\addlinespace[2pt]

$-$ & $+$
& \textsc{Affiliative \newline Positive}
& Aggressive joking among friends, affectionate blame, self-deprecation for attention, or criticism mixed with care and expectation.
& \textbf{Zh:} 你个猪，终于知道好好考一次了。 \newline
  \textbf{En:} You little pig, you finally learned to take an exam seriously. \\

\addlinespace[2pt]

$-$ & $0$
& \textsc{Reportive \newline Negative}
& News reporting, objective analysis, factual description, or neutral discussion of negative events.
& \textbf{Zh:} 事故造成多人受伤，现场交通一度中断。 \newline
  \textbf{En:} The accident injured several people, and traffic at the scene was temporarily suspended. \\

\addlinespace[2pt]

$-$ & $-$
& \textsc{Negative}
& Direct anger, blame, rejection, disappointment, anxiety, sadness, disgust, or dissatisfaction.
& \textbf{Zh:} 蠢货，给我滚出这里。 \newline
  \textbf{En:} You idiot, get out of here. \\

\bottomrule
\end{tabularx}

\caption{
Definitions and examples of the nine Affective Stances.
Each stance is determined by the composition of explicit affective signal $e_i$ and implicit affective tendency $h_i$.
}
\label{tab:stance_definitions}
\end{table*}
\end{CJK*}

\section{Formal Setup}
\label{app:ontology}

\subsection{Notation}
Let $D=(u_1,\ldots,u_T)$ be a dialogue. For a target utterance $u_t$, the context is $C_t=(u_{t-k},\ldots,u_{t-1})$ for a configurable window size $k$, or the full previous dialogue history when available.
Each benchmark instance is represented as $x_i=(C_i,u_i)$ and annotated with explicit affective signal, implicit affective tendency, Affective Stance, Pragmatic Intent, and Fine-grained Emotion.

\subsection{Well-Formedness}
An annotation is well formed only when the Affective Stance label matches the ordered pair $(y_i^{exp},y_i^{imp})$ under $\phi$.
Invalid stance combinations are rejected during export validation.
Pragmatic Intent and Fine-grained Emotion are not deterministic functions of the stance, but they must be justified by the same context--target pair and by the intermediate stance interpretation.

\section{Dataset Card}
\label{app:dataset_card}

\subsection{Intended Use}
The dataset is intended for research on Chinese discourse emotion understanding, Affective Stance Recognition, Pragmatic Intent Understanding, Fine-grained Emotion Classification, and robust affect-aware dialogue systems.
It is designed for evaluating how models infer unsaid affect from context rather than for reusing source-dataset sentiment or sarcasm annotations.

\subsection{Out-of-Scope Use}
The dataset should not be used to infer private mental states about real individuals, rank~\cite{liu2025learning,liu2026learning} users by emotional tendency, or make high-stakes decisions about employment, health, credit, or legal status.

\subsection{Source and License Notes}
The release records source identifiers so users can trace the text pool from which each instance was drawn.
Users should comply with the licenses and terms of the underlying source datasets, especially for sources with non-commercial restrictions.
CUE-Bench annotations are newly created and should not be interpreted as inherited labels from the original datasets.

\subsection{Recommended Reporting}
Papers using CUE-Bench should report the input setting, context window, prompt or model configuration, split version, Accuracy, macro-F1, weighted-F1, and whether any low-confidence or filtered examples are included.
For Task 1, Affective Stance Recognition, reports should include a confusion matrix over the nine stance categories, because this task corresponds to the Explicit-Implicit Stance Matrix.
For Task 2 and Task 3, reports should include class-wise F1 or per-class error analysis when space permits, since Pragmatic Intent and Fine-grained Emotion are long-tailed and more subjective.
When external data augmentation or additional annotation is used, papers should distinguish it clearly from the official CUE-Bench training, development, and test splits.

\section{Additional Evaluation Details}
\label{app:baselines}

\subsection{Input Formatting for LLM Evaluation}

Structured and reproducible evaluation protocols are also important in adjacent settings such as modular LLM adaptation, social-content prediction, retrieval, nearest-neighbor search, and action understanding~\cite{ICLR2026_LoRA-Mixer,ACMMM2026-Popularity,7REFINE,10STABLE,11EgoAction}.
We therefore report the input fields, prompting protocol, normalization rules, and oracle-conditioning settings used in CUE-Bench evaluation.

Each LLM input contains the dialogue context, the target utterance, 
task-specific label definitions, and a constrained output schema. 
For direct prompting settings, the model is asked to predict the target 
label directly from the given context and target utterance. For chain-based 
or matrix-guided prompting settings, the prompt specifies an intermediate 
reasoning order, but the intermediate fields are inferred by the model itself 
rather than provided as gold labels.

For Affective Stance Recognition, matrix-guided prompting asks the model to 
first infer the explicit affective signal and implicit affective tendency, 
and then map the ordered pair to one of the nine stance labels. For Pragmatic 
Intent and Fine-grained Emotion, matrix-guided prompting similarly requires 
the model to derive the relevant intermediate fields before predicting the 
final label. Gold intermediate labels are provided only in the 
oracle-conditioning ablation settings, where they are inserted into the prompt 
to test the contribution of each intermediate variable.

\subsection{Prompting Protocol}
Zero-shot prompting includes label definitions and a constrained output format.
Few-shot prompting adds demonstrations covering aligned affect, neutral-surface latent affect, and contrastive affect.
Free-form CoT prompting asks the model to explain its reasoning before giving the final label, while matrix-guided prompting fixes the reasoning order from explicit signal to implicit tendency, Affective Stance, Pragmatic Intent, and Fine-grained Emotion.
All prompts are evaluated with deterministic decoding when the API or model interface supports it.

\subsection{Prediction Normalization}
Model outputs are normalized before scoring.
We strip formatting artifacts, map aliases to canonical label names, and reject outputs that cannot be matched to the task label space.
When a response contains both rationale text and a final answer, only the final structured label field is used for metric computation.
This prevents verbose reasoning from being treated as additional labels.

\subsection{Oracle-conditioning Settings}
For the ablation study, gold intermediate labels are inserted into the prompt while the target label remains hidden.
Task I provides gold Affective Stance and predicts Pragmatic Intent.
Task II provides gold Affective Stance and predicts Fine-grained Emotion.
Task III provides both gold Affective Stance and gold Pragmatic Intent and predicts Fine-grained Emotion.





\end{document}